# Symbolic Neutrosophic Theory

| * | $I_1$ | $I_2$ | $I_3$ | $I_4$ | $I_5$ | $I_6$ |
|---|---|---|---|---|---|---|
| $I_1$ | $I_1$ | $I_1$ | $I_6$ | $I_6$ | $I_6$ | $I_6$ |
| $I_2$ | $I_1$ | $I_2$ | $I_3$ | $I_6$ | $I_3$ | $I_6$ |
| $I_3$ | $I_6$ | $I_3$ | $I_3$ | $I_6$ | $I_3$ | $I_6$ |
| $I_4$ | $I_6$ | $I_6$ | $I_6$ | $I_4$ | $I_4$ | $I_6$ |
| $I_5$ | $I_6$ | $I_3$ | $I_3$ | $I_4$ | $I_5$ | $I_6$ |
| $I_6$ | $I_6$ | $I_6$ | $I_6$ | $I_6$ | $I_6$ | $I_6$ |

Florentin Smarandache

Florentin Smarandache

# Symbolic Neutrosophic Theory

Florentin Smarandache

# Symbolic Neutrosophic Theory

**On the cover: Sub-Indeterminacies Multiplication Law Table**

# Contents

# Foreword

Symbolic (or Literal) Neutrosophic Theory is referring to the use of abstract symbols (i.e. the letters *T*, *I*, *F*, representing the neutrosophic components truth, indeterminacy, and respectively falsehood, or their refined components, represented by the indexed letters $T_j$, $I_k$, $F_l$) in neutrosophics. This book treats the neutrosophy, neutrosophic logic, neutrosophic set, and partially neutrosophic probability.

In *the first chapter*, we extend the dialectical triad **thesis-antithesis-synthesis** (dynamics of <A> and <antiA>, to get a synthesis) to the neutrosophic tetrad **thesis-antithesis-neutrothesis-neutrosynthesis** (dynamics of <A>, <antiA>, and <neutA>, in order to get a neutrosynthesis). We do this for better reflecting our world, since the neutralities between opposites play an important role. The *neutrosophic synthesis (neutrosynthesis)* is more refined that the dialectical synthesis. It carries on the unification and synthesis regarding the opposites and their neutrals too.

In *the second chapter,* we introduce for the first time the **neutrosophic system** and **neutrosophic dynamic system** that represent new perspectives in science. A neutrosophic system is a quasi- or $(t, i, f)$–classical system, in the sense that the neutrosophic system deals with quasi-terms/concepts/attributes, etc. [or $(t, i, f)$ -terms/ concepts/attributes], which are approximations of the classical terms/concepts/attributes, i.e. they are partially true/membership/probable ($t$%), partially indeterminate ($i$%), and partially false/nonmembership/improbable ($f$%), where $t, i, f$ are subsets of the unitary interval $[0, 1]$. {We recall that 'quasi' means relative(ly), approximate(ly), almost, near, partial(ly), etc. or mathematically 'quasi' means *(t,i,f)* in a neutrophic way.}

Thus we present in a neutrosophic (dynamic or not) system the $(t, i, f) - patterns,$ $(t, i, f)$ *-principles, (t,i,f)-laws, (t,i,f)-behavior, (t,i,f)-relationships, (t,i,f)-attractor* and *(t,i,f)-repellor, the thermodynamic (t,i,f)-equilibrium,* and so on.

In *the third chapter,* we introduce for the first time the notions of *Neutrosophic Axiom, Neutrosophic Deducibility, Neutrosophic Axiomatic System, Neutrosophic Deducibility and Neutrosophic Inference, Neutrosophic Proof, Neutrosophic Tauto-* *logies, Neutrosophic Quantifiers, Neutrosophic Propositional Logic, Neutrosophic Axiomatic Space, Degree of Contradiction (Dissimilarity) of Two Neutrosophic Axioms, and Neutrosophic Model.*

A class of neutrosophic implications is also introduced. A comparison between these innovatory neutrosophic notions and their corresponding classical notions is also made. Then, three concrete examples of neutrosophic axiomatic systems, describing the same neutrosophic geometrical model, are presented at the end of the chapter.

*The fourth chapter* is an improvement of our paper "(t, i, f)-Neutrosophic Structures" *[3: 1],* where we introduced for the first time a new type of structures, called *(t, i, f)-Neutrosophic Structures*, presented from a neutrosophic logic perspective, and we showed particular cases of such structures in geometry and in algebra.

In any field of knowledge, each structure is composed from two parts: a **space**, and a **set of axioms** (or **laws**) acting (governing) on it. If the space, or at least one of its axioms (laws), has some indeterminacy of the form *(t, i, f)* $\neq$ *(1, 0, 0),* that structure is a *(t, i, f)-Neutrosophic Structure*.

The *(t, i, f)-Neutrosophic Structures* [based on the components *t* = truth, *i* = numerical indeter-minacy, $f$ = falsehood] are different from the *Neutrosophic Algebraic Structures* [based on neutrosophic numbers of the form *a + bI,* where I = literal indeterminacy and $I^n = I$], that we rename as *I-Neutrosophic Algebraic Structures* (meaning algebraic structures based on indeterminacy "I" only). But we can combine both and obtain the *(t, i, f)-I-Neutrosophic Algebraic Structures*, i.e. algebraic structures based on neutrosophic numbers of the form *a+bI*, but also having indeterminacy of the form $(t, i, f) \neq (1, 0, 0)$ related to the structure space (elements which only partially belong to the space, or elements we know nothing if they belong to the space or not) or indeterminacy of the form $(t, i, f) \neq (1, 0, 0)$ related to at least one axiom (or law) acting on the structure space. Then we extend them to *Refined (t, i, f)- Refined I-Neutrosophic Algebraic Structures.*

In *the fifth chapter*, we make a short history of: the *neutrosophic set, neutrosophic numerical components and neutrosophic literal components, neutrosophic numbers, neutrosophic intervals, neutrosophic dual number, neutrosophic special dual number, neutrosophic special quasi dual number, neutrosophic quaternion number, neutrosophic octonion number, neutrosophic linguistic number, neutrosophic linguistic interval-style number,* *neutrosophic hypercomplex numbers of dimension* n*, and elemen-tary neutrosophic algebraic structures*. Afterwards, their generalizations to *refined neutrosophic set*, respectively *refined neutrosophic numerical and literal components*, then *refined neutrosophic numbers and refined neutrosophic algebraic structures*, and *set-style neutrosophic numbers*.

The aim of this chapter is to construct examples of splitting the literal indeterminacy *(I)* into literal sub-indeterminacies $(I_1, I_2, \ldots, I_r)$, and to define a *multiplication law* of these literal sub-indeterminacies in order to be able to build refined *I-neutrosophic algebraic structures*. Also, we give examples of splitting the numerical indeterminacy *(i)* into numerical sub-indeterminacies, and examples of splitting neutrosophic numerical components into neutrosophic numerical sub-components.

In *the sixth chapter*, we define for the first time three *neutrosophic actions* and their properties. We then introduce the *prevalence order* on $\{T, I, F\}$ with respect to a given neutrosophic operator "*o*", which may be subjective - as defined by the neutrosophic experts. And the *refinement of neutrosophic entities* <A>, <neutA>, and <antiA>.

Then we extend the classical logical operators to *neutrosophic literal (symbolic) logical operators* and to *refined literal (symbolic) logical operators*, and we define the *refinement neutrosophic literal (symbolic) space*.

In *the seventh chapter*, we introduce for the first time the *neutrosophic quadruple numbers* (of the form $a + bT + cI + dF$) and the *refined neutrosophic quadruple numbers*.

Then we define an *absorbance law*, based on a *prevalence order*, both of them in order to multiply the neutrosophic components $T, I, F$ or their sub-components $T_j, I_k, F_l$ and thus to construct the *multiplication of neutrosophic quadruple numbers*.

## Note.

Parts of this book have been partially published before as it follows:

- *Thesis-Antithesis-Neutrothesis, and Neutrosynthesis*, in Neutrosophic Sets and Systems, 64-67, Vol. 8, 2015.
- *Neutrosophic Axiomatic System*, in Critical Review, Center for Mathematics of Uncertainty, Creighton University, Omaha, NE, USA, Vol. X, 5-28, 2015.
- *(T, I, F)-Neutrosophic Structures*, in Journal of Advance in Mathematical Science, International Institute For Universal Research (IIUR), Vol. 1, Issue 1, 70-80, 2015;

  and in Proceedings of ICMERA 2015, Bucharest, Romania, 2015; also presented at the Annual Symposium of the Institute of Solid Mechanics, SISOM 2015, Robotics and Mechatronics. Special Session and Work Shop on VIPRO Platform and RABOR Rescue Robots, Romanian Academy, Bucharest, 21-22 May 2015.
- *Refined Literal Indeterminacy and the Multiplication Law of Subindeterminacies*, in Neutrosophic Sets and Systems, 58-63, Vol. 9, 2015.
- *Neutrosophic Quadruple Numbers, Refined Neutrosophic Quadruple Numbers, Absorbance Law, and the Multiplication of Neutrosophic Quadruple Numbers*, in Neutrosophic Sets and Systems, Vol. 10, 2015 [under print].

- *Neutrosophic Actions, Prevalence Order, Refinement of Neutrosophic Entities, and Neutrosophic Literal Logical Operators*, in Critical Review, Center for Mathematics of Uncertainty, Creighton University, Omaha, NE, USA, Vol. XI, 2015 [under print].
- Neutrosophic Systems and Neutrosophic Dynamic Systems, in Critical Review, Center for Mathematics of Uncertainty, Creighton University, Omaha, NE, USA, Vol. XI, 2015 [under print].

# 1 Thesis-Antithesis-Neutrothesis, and Neutrosynthesis

## 1.1 Abstract.

In this chapter we extend the dialectical triad **thesis-antithesis-synthesis** (dynamics of <A> and <antiA>, to get a synthesis) to the neutrosophic tetrad **thesis-antithesis-neutrothesis-neutrosynthesis** (dynamics of <A>, <antiA>, and <neutA>, in order to get a neutrosynthesis). We do this for better reflecting our world, since the neutralities between opposites play an important role. The *neutrosophic synthesis (neutrosynthesis)* is more refined that the dialectical synthesis. It carries on the unification and synthesis regarding the opposites and their neutrals too.

## 1.2 Introduction.

In neutrosophy, <A>, <antiA>, and <neutA> combined two by two, and also all three of them together form the NeutroSynthesis. Neutrosophy establishes the universal relations between <A>, <antiA>, and <neutA>.

<A> is the thesis, <antiA> the antithesis, and <neutA> the neutrothesis (neither <A> nor <antiA>, but the neutrality in between them).

In the neutrosophic notation, <nonA> (not <A>, outside of <A>) is the union of <antiA> and <neutA>.

<neutA> may be from no middle (*excluded middle*), to one middle (*included middle*), to many finite discrete middles (*finite multiple included-middles*), and to an infinitude of discrete or continuous middles (*infinite multiple included-middles*) [for example, as in color for the last one, let's say between black and white there is an infinite spectrum of middle/intermediate colors].

## 1.3 Thesis, Antithesis, Synthesis.

The classical reasoning development about evidences, popularly known as thesis-antithesis-synthesis from dialectics, was attributed to the renowned philosopher Georg Wilhelm Friedrich Hegel (1770-1831) and later it was used by Karl Marx (1818-1883) and Friedrich Engels (1820-1895). About thesis and antithesis have also written Immanuel Kant (1724-1804), Johann Gottlieb Fichte (1762-1814), and Thomas Schelling (born 1921).

In ancient Chinese philosophy the opposites *yin* [feminine, the moon] and *yang* [masculine, the sun] were considered complementary.

## 1.4 Thesis, Antithesis, Neutrothesis, Neutrosynthesis.

Neutrosophy is a generalization of dialectics (which is based on contradictions only, <A> and <antiA>), because neutrosophy is based on contradictions and on the neutralities between them (<A>, <antiA>, and <neutA>). Therefore, the dialectical triad **thesis-antithesis-synthesis** is extended to the neutrosophic tetrad **thesis-antithesis-neutrothesis-neutrosynthesis**. We do this not for the sake of generalization, but for better reflecting our world. A neutrosophic synthesis (neutrosynthesis) is more refined that the dialectical synthesis. It carries on the unification and synthesis regarding the opposites and their neutrals too.

## 1.5 Neutrosophic Dynamicity.

We have extended in [1] the *Principle of Dynamic Opposition* [opposition between <A> and <antiA>] to the ***Principle of Dynamic Neutropposition*** [which means oppositions among <A>, <antiA>, and <neutA>].

Etymologically "neutropposition" means "neutrosophic opposition".

This reasoning style is not a neutrosophic scheme, but it is based on reality, because if an idea (or notion) <A> arises, then multiple versions of this idea are spread out, let's denote them by <A>$_1$, <A>$_2$, …, <A>$_m$. Afterwards, the opposites (in a smaller or higher degree) ideas are born, as reactions to <A> and its versions <A>$_i$. Let's denote these versions of opposites by <antiA>$_1$, <antiA>$_2$, …, <antiA>$_n$. The neutrality <neutA> between these contradictories ideas may embrace various forms, let's denote them by <neutA>$_1$, <neutA>$_2$, …, <neutA>$_p$, where *m, n, p* are integers greater than or equal to *1*.

In general, for each <A> there may be corresponding many <antiA>'s and many <neutA>'s. Also, each <A> may be interpreted in many different versions of <A>'s too.

Neutrosophic Dynamicity means the interactions among all these multi-versions of <A>'s with their multi-<antiA>'s and their multi-<neutA>'s, which will result in a new thesis, let's call it <A'> at a superior level. And a new cycle of <A'>, <antiA'>, and <neutA'> restarts its neutrosophic dynamicity.

## 1.6 Practical Example.

Let's say <A> is a country that goes to war with another country, which can be named <antiA> since it is antagonistic to the first country. But many neutral countries <neutA> can interfere, either supporting or aggressing one of them, in a smaller or bigger degree. Other neutral countries <neutA> can still remain neutral in this war. Yet, there is a continuous dynamicity between the three categories (<A>, <antiA>, <neutA.), for countries changing sides (moving from a coalition to another coalition), or simply retreating from any coalition.

In our easy example, we only wanted to emphasize the fact that **<neutA> plays a role in the conflict between the opposites <A> and <antiA>**, role which was ignored by dialectics.

So, the dialectical synthesis is extended to a neutrosophic synthesis, called neutrosynthesis, which combines thesis, antithesis, and neutrothesis.

## 1.7 Theoretical Example.

Suppose <A> is a philosophical school, and its opposite philosophical school is <antiA>. In the dispute between <A> and <antiA>, philosophers from the two contradictory groups may bring arguments against the other philosophical school from various neutral philosophical schools' ideas (<neutA>, which were neither for <A> nor <antiA>) as well.

## 1.8 Acknowledgement.

The author would like to thank Mr. Mumtaz Ali, from Quaid-i-Azam University, Islamabad, Pakistan, for his comments on the paper.

## 1.9 References.

# 2 Neutrosophic Systems and Neutrosophic Dynamic Systems

## 2.1 Abstract.

In this chapter, we introduce for the first time the **neutrosophic system** and **neutrosophic dynamic system** that represent new per-spectives in science. A neutrosophic system is a quasi- or $(t, i, f)$–classical system, in the sense that the neutrosophic system deals with quasi-terms/concepts/attributes, etc. [or $(t, i, f)$-terms/ concepts/attributes], which are approximations of the classical terms/concepts/attributes, i.e. they are partially true/membership/probable ( $t\%$ ), partially indeterminate ($i\%$), and partially false/nonmembership/improbable ($f\%$), where $t, i, f$ are subsets of the unitary interval $[0, 1]$. {We recall that 'quasi' means relative(ly), approximate(ly), almost, near, partial(ly), etc. or mathematically 'quasi' means *(t,i,f)* in a neutrophic way.}

## 2.1 Introduction.

A **system** $\mathcal{S}$ in general is composed from a **space** $\mathcal{M}$, together with its **elements** (concepts) $\{e_j\}$, $j \in \theta$, and the **relationships** $\{\mathcal{R}_k\}$, $k \in \psi$, between them, where $\theta$ and $\psi$ are countable or uncountable index sets.

For a **closed system**, the space and its elements do not interact with the environment.

For an **open set**, the space or its elements interact with the environment.

## 2.2 Definition of the neutrosophic system.

A system is called **neutrosophic system** if at least one of the following occur:

1. The space contains some indeterminacy.
2. At least one of its elements $x$ has some indeterminacy (it is not well-defined or not well-known).
3. At least one of its elements $x$ does not 100% belong to the space; we say $x(t, i, f) \in \mathcal{M}$, with $(t, i, f) \neq (1, 0, 0)$.
4. At least one of the relationships $\mathcal{R}_o$ between the elements of $\mathcal{M}$ is not 100% well-defined (or well-known); we say $\mathcal{R}_o(t, i, f) \in \mathcal{S}$, with $(t, i, f) \neq (1, 0, 0)$.

5. For an open system, at least one $[\mathcal{R}_E(t,i,f)]$ of the system's interactions relationships with the environment has some indeterminacy, or it is not well-defined, or not well-known, with $(t,i,f) \neq (1,0,0)$.

### 2.2.1 Classical system as particular case of neutrosophic system.

By language abuse, a classical system is a neutrosophic system with indeterminacy zero (no indeterminacy) at all system's levels.

### 2.2.2 World systems are mostly neutrosophic.

In our opinion, most of our world systems are neutrosophic systems, not classical systems, and the dynamicity of the systems is neutrosophic, not classical.

Maybe the mechanical and electronical systems could have a better chance to be classical systems.

## 2.3 A simple example of neutrosophic system.

Let's consider a university campus Coronado as a whole neutrosophic system $\mathcal{S}$, whose space is a prism having a base the campus land and the altitude such that the prism encloses all campus' buildings, towers, observatories, etc.

The elements of the space are people (administration, faculty, staff, and students) and objects (buildings, vehicles, computers, boards, tables, chairs, etc.).

A part of the campus land is unused. The campus administration has not decided yet what to do with it: either to build a laboratory on it, or to sell it. This is an indeterminate part of the space.

Suppose that a staff (John, from the office of Human Resources) has been fired by the campus director for misconduct. But, according to his co-workers, John was not guilty for anything wrong doing. So, John sues the campus. At this point, we do not know if John belongs to the campus, or not. John's appurtenance to the campus is indeterminate.

Assume the faculty norm of teaching is four courses per semester. But some faculty are part-timers, therefore they teach less number of courses. If an instructor teaches only one class per semester, he belongs to the campus only partially (25%), if he teaches two classes he belongs to the campus 50%, and if he teaches three courses he belongs to the campus 75%. We may write:

Joe $(0.25, 0, 0.75) \in \mathcal{S}$

George $(0.50, 0, 0.50) \in \mathcal{S}$

and Thom $(0.75, 0.10, 0.25) \in \mathcal{S}$.

Thom has some indeterminacy (0.10) with respect to his work in the campus: it is possible that he might do some administrative work for the campus (but we don't know).

The faculty that are full-time (teaching four courses per semester) may also do overload. Suppose that Laura teaches five courses per semester, therefore Laura $(1.25, 0, 0) \in \mathcal{S}$.

In neutrosophic logic/set/probability it's possible to have the sum of components $(t, i, f)$ different from 1:

$t + i + f > 1$, for paraconsistent (conflicting) information;

$t + i + f = 1$, for complete information;

$t + i + f < 1$, for incomplete information.

Also, there are staff that work only ½ norm for the campus, and many students take fewer classes or more classes than the required full-time norm. Therefore, they belong to the campus Coronado in a percentage different from 100%.

About the objects, suppose that 50 calculators were brought from IBM for one semester only as part of IBM's promotion of their new products. Therefore, these calculators only partially and temporarily belong to the campus.

Thus, not all elements (people or objects) entirely belong to this system, there exist many $e_j(t, i, f) \in \mathcal{S}$, with $(t, i, f) \neq (1, 0, 0)$.

Now, let's take into consideration the relationships. A professor, Frank, may agree with the campus dean with respect to a dean's decision, may disagree with respect to the dean's other decision, or may be ignorant with respect to the dean's various decisions. So, the relationship between Frank and the dean may be, for example:

$$\text{Frank} \xrightarrow{\text{agreement }(0.5,0.2,0.3)} \text{dean, i. e. not } (1, 0, 0)\ \mathit{agreement}.$$

This campus, as an open system, co-operates with one Research Laboratory from Nevada, pending some funds allocated by the government to the campus.

Therefore, the relationship (research co-operation) between campus Coronado and the Nevada Research Laboratory is indeterminate at this moment.

## 2.4 Neutrosophic patterns.

In a neutrosophic system, we may study or discover, in general, **neutrosophic patterns**, i.e. quasi-patterns, approximated patterns, not totally working; we say: $(t, i, f)$ -patterns, i.e. $t\%$ true, $i\%$ indeterminate, and $f$% false, and elucidate $(t, i, f)$-principles.

The neutrosophic system, through feedback or partial feedback, is $(t, i, f)$ – self-correcting, and $(t, i, f)$-self-organizing.

## 2.5 Neutrosophic holism.

From a holistic point of view, the sum of parts of a system may be:

1. Smaller than the whole (when the interactions between parts are unsatisfactory);
2. Equals to the whole (when the interactions between parts are satisfactory);
3. Greater than the whole (when the interactions between parts are super-satisfactory).

The more interactions (interdependance, transdependance, hyperdependance) between parts, the more complex a system is.

We have positive, neutral, and negative interactions between parts.

Actually, an interaction between the parts has a degree of positiveness, degree of neutrality, and degree of negativeness.

And these interactions are dynamic, meaning that their degrees of positiveness/ neutrality/negativity change in time.

They may be partially absolute and partially relative.

## 2.6 Neutrosophic model.

In order to model such systems, we need a neutrosophic (approximate, partial, incomplete, imperfect) model that would discover the approximate system properties.

## 2.7 Neutrosophic successful system.

A neutrosophic successful system is a system that is successful with respect to some goals, and partially successful or failing with respect to other goals.

The adaptivity, self-organization, self-reproducing, self-learning, reiteration, recursivity, relationism, complexity and other attributes of a classical system are extended to $(t, i, f)$-attributes in the neutrosophic system.

## 2.8 $(t, i, f)$-attribute.

A $(\boldsymbol{t}, \boldsymbol{i}, \boldsymbol{f})$**-attribute** means an attribute that is $t$% true (or probable), $i$% indeterminate (with respect to the true/probable and false/improbable), and $f\%$ false/improbable - where $t, i, f$ are subsets of the unitary interval $[0, 1]$.

For example, considering the subsets reduced to single numbers, if a neutrosophic system is (0.7, 0.2, 0.3)-adaptable, it means that the system is 70% adaptable, 20% indeterminate regarding adaptability, and 30% inadaptable; we may receive the informations for each attribute phase from different independent sources, that's why the sum of the neutrosophic components is not necessarily 1.

## 2.9 Neutrosophic dynamics.

While classical dynamics was beset by dialectics, which brought together an entity $\langle A \rangle$ and its opposite $\langle antiA \rangle$, the neutrosophic dynamics is beset by tri-alectics, which brings together an entity $\langle A \rangle$ with its opposite $\langle antiA \rangle$ and their neutrality $\langle neutA \rangle$. Instead of duality as in dialectics, we have tri-alities in our world.

Dialectics failed to take into consideration the neutrality between opposites, since the neutrality partially influences both opposites.

Instead of unifying the opposites, the neutrosophic dynamics unifies the triad $\langle A \rangle, \langle antiA \rangle, \langle neutA \rangle$.

Instead of coupling with continuity as the classical dynamics promise, one has “tripling” with continuity and discontinuity altogether.

All neutrosophic dynamic system’s components are interacted in a certain degree, repelling in another degree, and neutral (no interaction) in a different degree.

They comprise the systems whose equilibrium is the disechilibrium - systems that are continuously changing.

The internal structure of the neutrosophic system may increase in complexity and interconnections, or may degrade during the time.

A neutrosophic system is characterized by potential, impotential, and indeterminate developmental outcome, each one of these three in a specific degree.

## 2.10 Neutrosophic behavior gradient.

In a neutrosophic system, we talk also about **neutrosophic structure**, which is actually a quasi-structure or structure which manifests into a certain degree; which influences the **neutrosophic behavior gradient**, that similarly is a behavior quasi-gradient - partially determined by quasi-stimulative effects; one has: discrete systems, continuous systems, hybrid (discrete and continuous) systems.

## 2.11 Neutrosophic interactions.

Neutrosophic interactions in the system have the form:

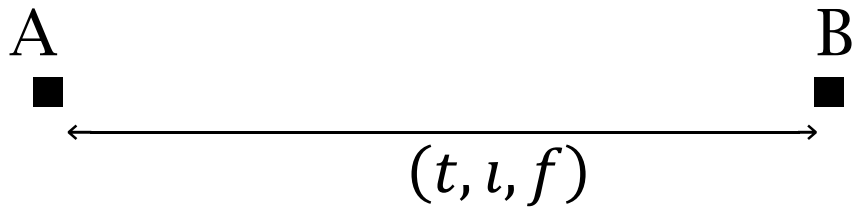

Neutrosophic self-organization is a quasi-self-organization.

The system's **neutrosophic intelligence** sets into the neutrosophic patterns formed within the system's elements.

We have a neutrosophic causality between event $E_1$, that triggers event $E_2$, and so on. And similarly, neutrosophic structure $S_1$ (which is an approximate, not clearly know structure) causes the system to turn on neutrosophic structure $S_2$, and so on. A neutrosophic system has different levels of self-organizations.

## 2.12 Potentiality/impotentiality/indeterminacy.

Each neutrosophic system has a potentiality/impotentiality/indeterminacy to attain a certain state/stage; we mostly mention herein about the transition from a **quasi-pattern** to another quasi-pattern. A neutrosophic open system is always transacting with the environment; since always the change is needed.

A neutrosophic system is always oscilating between stability, instability, and ambiguity (indeterminacy).

Analysis, synthesis, and neutrosynthesis of existing data are done by the neutrosophic system. They are based on system's principles, anti-principles, and nonprinciples.

## 2.13 Neutrosophic synergy.

The **Neutrosophic Synergy** is referred to partially joined work or partially combined forces, since the participating forces may cooperate in a degree ($t$), may be antagonist in another degree ($f$), and may have a neutral interest in joint work in a different degree ($i$).

## 2.14 Neutrosophic complexity.

The neutrosophic complex systems produce neutrosophic complex patterns. These patterns result according to the neutrosophic relationships among system's parts. They are well described by the neutrosophic cognitive maps (*NCM*), neutrosophic relational maps (*NRM*), and neutrosophic relational equations (NRE), all introduced by

W. B. Vasanttha Kandasamy and F. Smarandache in 2003-2004.

The neutrosophic systems represent a new perspective in science. They deal with quasi-terms [or $(t,i,f)$ -terms], quasi-concepts [or $(t,i,f)$ -concepts], and quasi-attributes [or $(t,i,f)$-attributes], which are approximations of the terms, concepts, attributes, etc., i.e. they are partially true ( $t\%$ ), partially indeterminate ($i\%$), and partially false ($f\%$).

Alike in neutrosophy, where there are interactions between $\langle A\rangle$, $\langle neutA\rangle$, and $\langle antiA\rangle$, where $\langle A\rangle$ is an entity, a system is frequently in one of these general states: equilibrium, indeterminacy (neither equilibrium, nor disequilibrium), and disequilibrium.

They form a **neutrosophic complexity** with neutrosophically ordered patterns. A neutrosophic order is a quasi or approximate order, which is described by a neutrosophic formalism.

The parts all together are partially homogeneous, partially heterogeneous, and they may combine in finitely and infinitely ways.

## 2.15 Neutrosophic processes.

The neutrosophic patterns formed are also dynamic, changing in time and space. They are similar, dissimilar, and indeterminate (unknown,

hidden, vague, incomplete) processes among the parts. They are called neutrosophic processes.

## 2.16 Neutrosophic system behavior.

The neutrosophic system's functionality and behavior are, therefore, coherent, incoherent, and imprevisible (indeterminate). It moves, at a given level, from a neutrosophic simplicity to a neutrosophic complexity, whch becomes neutrosophic simplicity at the next level. And so on.

Ambiguity (indeterminacy) at a level propagates at the next level.

## 2.17 Classical systems.

Although the biologist Bertalanffy is considered the father of *general system theory* since 1940, it has been found out that the conceptual portion of the system theory was published by Alexander Bogdanov between 1912-1917 in his three volumes of *Tectology*.

## 2.18 Classical open systems.

A classical open system, in general, cannot be totally deterministic, if the environment is not totally deterministic itself.

Change in energy or in momentum makes a classical system to move from thermodynamic equilibrium to nonequilibrium or reciprocally.

Open classical systems, by infusion of outside energy, may get an unexpected spontaneous structure.

## 2.19 Deneutrosophication.

In a neutrosophic system, besides the degrees of freedom, one also talk about the degree (grade) of indeterminacy. Indeterminacy can be described by a variable.

Surely, the degrees of freedom should be condensed, and the indetermination reduced (the last action is called “deneutrosophication”).

The neutrosophic system has a multi-indeterminate behavior. A neutrosophic operator of many variables, including the variable representing indeterminacy, can approximate and semi-predict the system’s behavior.

## 2.10 From classical to neutrosophic systems.

Of course, in a bigger or more degree, one can consider the *neutrosophic cybernetic system* (quasi or approximate control mechanism, quasi information processing, and quasi information reaction), and similarly the *neutrosophic chaos theory*, *neutrosophic catastrophe theory*, or *neutrosophic complexity theory*.

In general, when passing from a classical system $\mathcal{S}_c$ in a given field of knowledge $\mathcal{F}$ to a corresponding neutrosophic system $\mathcal{S}_N$ in the same field of knowledge $\mathcal{F}$, one relaxes the restrictions about the system's space, elements, and relationships, i.e. these components of the system (space, elements, relationships) may contain indeterminacy, may be partially (or totally) unknown (or vague, incomplete, contradictory), may only partially belong to the system; they are approximate, quasi.

Scientifically, we write:

$$\mathcal{S}_N = (t, i, f) - \mathcal{S}_c, \qquad (1)$$

and we read: a neutrosophic system is a $(t, i, f)$-classical system. As mapping, between the neutrosophic algebraic structure systems, we have defined *neutrosophic isomorphism*.

## 2.21 Neutrosophic dynamic system.

The behavior of a neutrosophic dynamic system is chaotic from a classical point of view. Instead of *fixed points*, as in classical dynamic systems, one deals with fixed regions (i.e. neigborhoods of fixed points), as approximate values of the neutrosophic variables [we recall that a neutrosophic variable is, in general, represented by a thick curve – alike a neutrosophic (thick) function].

There may be several fixed regions that are *attractive regions* in the sense that the neutrosophic system converges towards these regions if it starts out in a nearby neutrosophic state.

And similarly, instead of periodic points, as in classical dynamic systems, one has *periodic regions*, which are neutrosophic states where the neutrosophic system repeats from time to time.

If two or more periodic regions are non-disjoint (as in a classical dynamic system, where the fixed points lie in the system space too close to each other, such that their corresponding neighborhoods intersect), one gets *double periodic region*, *triple periodic region*:

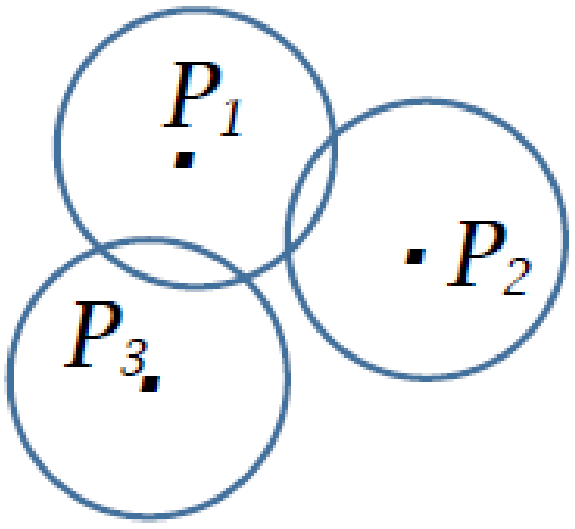

*Fig. 1*

and so on: *n-uple periodic region, for* $n \geq 2$.

In a simple/double/triple/…/n-uple periodic region the neutrosophic system is fluctuating/ oscilating from a point to another point.

The smaller is a fixed region, the better is the accuracy.

## 2.22 Neutrosophic cognitive science.

In the Neutrosophic Cognitive Science, the Indeterminacy "I" led to the definition of the *Neutrosophic Graphs* (graphs which have: either at least one indeterminate edge, or at least one indeterminate vertex, or both some indeterminate edge and some indeterminate vertex), and *Neutrosophic Trees* (trees which have: either at least one indeterminate edge, or at least one indeterminate vertex, or both some indeterminate edge and some indeterminate vertex), that have many applications in social sciences.

Another type of neutrosophic graph is when at least one edge has a neutrosophic $(t, i, f)$ truth-value.

As a consequence, the Neutrosophic Cognitive Maps (Vasantha & Smarandache, 2003) and *Neutrosophic Relational Maps* (Vasantha & Smarandache, 2004) are generalizations of fuzzy cognitive maps and respectively fuzzy relational maps,

*Neutrosophic Relational Equations* (Vasantha & Smarandache, 2004), *Neutrosophic Relational Data* (Wang, Smarandache, Sunderraman, Rogatko - 2008), etc.

A *Neutrosophic Cognitive Map* is a neutrosophic directed graph with concepts like policies, events etc. as vertices, and causalities or indeterminates as edges. It represents the causal relationship between concepts.

An edge is said indeterminate if we don't know if it is any relationship between the vertices it connects, or for a directed graph we don't know if it is a directly or inversely proportional relationship. We may write for such edge that $(t, i, f) = (0, 1, 0)$.

A vertex is indeterminate if we don't know what kind of vertex it is since we have incomplete information. We may write for such vertex that $(t, i, f) = (0, 1, 0)$.

Example of Neutrosophic Graph (edges $V_1V_3$, $V_1V_5$, $V_2V_3$ are indeterminate and they are drawn as dotted):

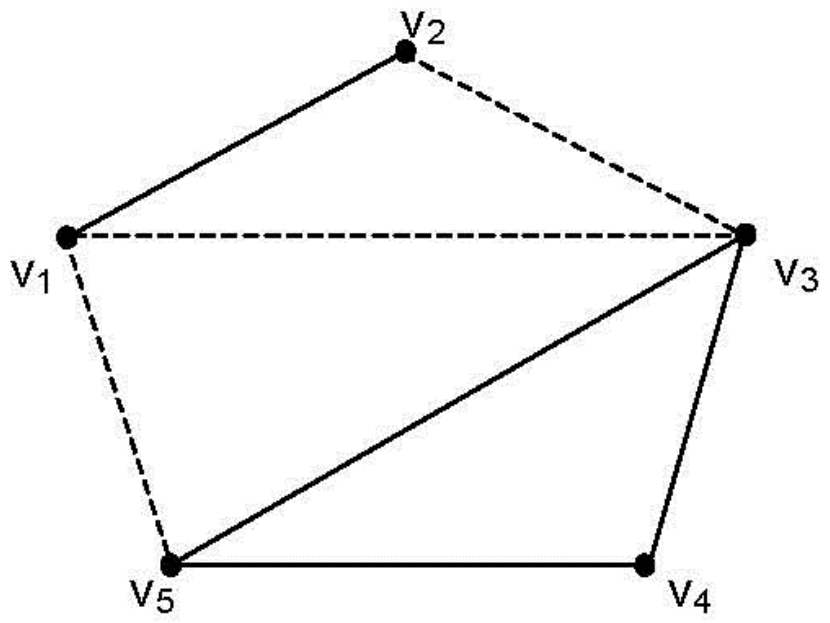

*Fig. 2*

and its neutrosophic adjacency matrix is:

$$\begin{bmatrix} 0 & 1 & I & 0 & I \\ 1 & 0 & I & 0 & 0 \\ I & I & 0 & 1 & 1 \\ 0 & 0 & 1 & 0 & 1 \\ I & 0 & 1 & 1 & 0 \end{bmatrix}$$

The edges mean: 0 = no connection between vertices, 1 = connection between vertices, I = indeterminate connection (not known if it is, or if it is not).

Such notions are not used in the fuzzy theory.

Let's give an example of Neutrosophic Cognitive Map (NCM), which is a generalization of the Fuzzy Cognitive Maps.

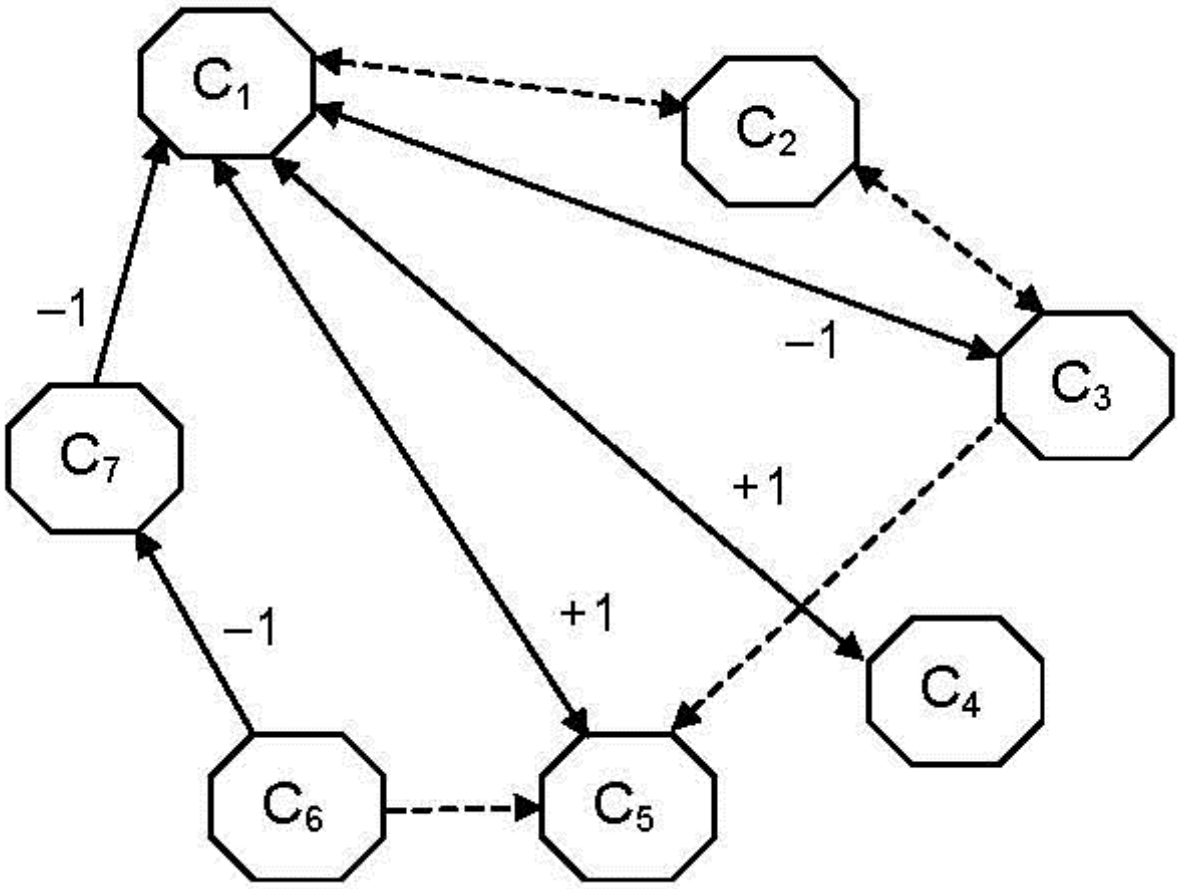

*Fig. 3*

We take the following vertices:

C1 - Child Labor

C2 - Political Leaders

C3 - Good Teachers

C4 - Poverty

C5 - Industrialists

C6 - Public practicing/encouraging Child Labor

C7 - Good Non-Governmental Organizations (NGOs)

The corresponding neutrosophic adjacency matrix related to this neutrosophic cognitive map is:

$$\begin{bmatrix} 0 & I & -1 & 1 & 1 & 0 & 0 \\ I & 0 & I & 0 & 0 & 0 & 0 \\ -1 & I & 0 & 0 & I & 0 & 0 \\ 1 & 0 & 0 & 0 & 0 & 0 & 0 \\ 1 & 0 & 0 & 0 & 0 & 0 & 0 \\ 0 & 0 & 0 & 0 & I & 0 & -1 \\ -1 & 0 & 0 & 0 & 0 & 0 & 0 \end{bmatrix}$$

The edges mean: 0 = no connection between vertices, 1 = directly proportional connection, -1 = inversely proportionally connection, and I = indeterminate connection (not knowing what kind of relationship is between the vertices that the edge connects).

Now, we give another type of neutrosophic graphs (and trees): An edge of a graph, let's say from *A* to *B* (i.e. how *A* influences *B*), may have a neutrosophic value *(t, i, f),* where t means the positive influence of *A* on *B*, i means the indeterminate/ neutral influence of *A* on *B*, and f means the negative influence of *A* on *B*.

Then, if we have, let's say: $A{-}{>}B{-}{>}C$ such that $A{-}{>}B$ has the neutrosophic value $(t_1, i_1, f_1)$

and $B -> C$ has the neutrosophic value $(t_2, i_2, f_2)$, then $A -> C$ has the neutrosophic value $(t_1, i_1, f_1)/\backslash(t_2, i_2. f_2)$, where /\ is the $AND_N$ neutrosophic operator.

Also, again a different type of graph: we consider a vertex A as: *t%* belonging/membership to the graph, *i%* indeterminate membership to the graph, and *f%* nonmembership to the graph.

Finally, one may consider any of the previous types of graphs (or trees) put together.

## 2.23 $(t, i, f)$-qualitative behavior.

We normally study in a neutrosophic dynamic system its long-term $(t, i, f)$ *-qualitative behavior*, i.e. degree of behavior's good quality (*t*), degree of behavior's indeterminate (unclear) quality (*i*), and degree of behavior's bad quality (*f*).

The questions arise: will the neutrosophic system fluctuate in a fixed region (considered as a neutrosophic steady state of the system)? Will the fluctuation be smooth or sharp? Will the fixed region be large (hence less accuracy) or small (hence bigger accuracy)? How many periodic regions does the neutrosophic system has? Do any of them intersect [i.e. does the neutrosophic system has some *n*-uple periodic regions (for $n \geq 2$), and for how many]?

## 2.24 Neutrosophic state.

The more indeterminacy a neutrosophic system has, the more chaotic it is from the classical point of view. A neutrosophic lineal dynamic system still has a degree of chaotic behavior. A collection of numerical sets determines a *neutrosophic state*, while a classical state is determined by a collection of numbers.

## 2.25 Neutrosophic evolution rule.

The *neutrosophic evolution rule* decribes the set of neutrosophic states where the future state (that follows from a given current state) belongs to.

If the set of neutrosophic states, that the next neutrosophic state will be in, is known, we have a *quasi-deterministic* neutrosophic evolution rule, otherwise the neutrosophic evolution rule is called *quasi-stochastic*.

## 2.26 Neutrosophic chaos.

As an alternative to the classical Chaos Theory, we have the Neutrosophic Chaos Theory, which is highly sensitive to indeterminacy; we mean that small change in the neutrosophic system's initial indeterminacy produces huge perturbations of the neutrosophic system's behavior.

## 2.27 Time quasi-delays and quasi-feedback thick-loops.

Similarly, the difficulties in modelling and simulating a *Neutrosophic Complex System* (also called *Science of Neutrosophic Complexity*) reside in its degree of indeterminacy at each system's level.

In order to understand the Neutrosophic System Dynamics, one studies the system's *time quasi-delays* and internal *quasi-feedback thick-loops* (that are similar to thick functions ad thick curves defined in the neutrosophic precalculus and neutrosophic calculus).

The system may oscillate from linearity to nonlinearity, depending on the neutrosophic time function.

## 2.28 Semi-open semi-closed system.

Almost all systems are open (exchanging energy with the environment).

But, in theory and in laboratory, one may consider closed systems (completely isolated from the environment); such systems can oscillate between closed and open (when they are cut from the environment, or put back in contact with the environment respectively).

Therefore, between open systems and closed systems, there also is a semi-open semi-closed system.

## 2.29 Neutrosophic system's development.

The system's self-learning, self-adapting, self-conscienting, self-developing are parts of the system's dynamicity and the way it moves from a state to another state – as a response to the system internal or external conditions. They are constituents of system's behavior.

The more developed is a neutrosophic system, the more complex it becomes. System's development depends on the internal and external interactions (relationships) as well.

Alike classical systems, the neutrosophic system shifts from a quasi-developmental level to another. Inherent fluctuations are characteristic to neutrosophic complex systems. Around the quasi-steady states, the fluctuations in a neutrosophic system becomes its sources of new quasi-development and quasi-behavior.

In general, a neutrosophic system shows a nonlinear response to its initial conditions.

The environment of a neutrosophic system may also be neutrosophic (i.e. having some indeterminacy).

## 2.30 Dynamic dimensions of neutrosophic systems.

There may be neutrosophic systems whose spaces have *dynamic dimensions*, i.e. their dimensions change upon the time variable.

Neutrosophic Dimension of a space has the form $(t, i, f)$, where we are *t%* sure about the real dimension of the space, *i%* indeterminate about the real dimension of the space, and *f%* unsure about the real dimension of the space.

## 2.31 Noise in a neutrosophic system.

A neutrosophic system's noise is part of the system's indeterminacy. A system's pattern may evolve or dissolve over time, as in a classical system.

## 2.32 Quasi-stability.

A neutrosophic system has a degree of stability, degree of indeterminacy referring to its stability, and degree of instability.

Similarly, it has a degree of change, degree of indeterminate change, and degree of non-change at any point in time.

*Quasi-stability* of a neutrosophic system is its partial resistance to change.

## 2.33 $(t, i, f)$-attractors.

Neutrosophic system's quasi-stability is also dependant on the $(t, i, f)$-attractor, which $t\%$ attracts, $i\%$ its attraction is indeterminate, and $f\%$ rejects. Or we may say that the neutrosophic system $(t\%, i\%, f\%)$-prefers to reside in a such neutrosophic attractor.

Quasi-stability in a neutrosophic system responds to quasi-perturbations.

When $(t, i, f) \to (1, 0, 0)$ the quasi-attractors tend to become stable, but if $(t, i, f) \to (0, i, f)$, they tend to become unstable.

Most neutrosophic system are very chaotic and possess many quasi-attractors and anomalous quasi-patterns. The degree of freedom in a neutrosophic complex system increase and get more intricate due to the type of indeterminacies that are specific to that system. For example, the classical system's noise is a sort of indeterminacy.

Various neutrosophic subsystems are assembled into a neutrosophic complex system.

## 2.34 $(t, i, f)$-repellors.

Besides attractors, there are systems that have *repellors*, i.e. states where the system avoids residing. The neutrosophic systems have quasi-repellors, or $(t, i, f)$ *-repellors*, i.e. states where the neutrosophic system partialy avoid residing.

## 2.35 Neutrosophic probability of the system's states.

In any (classical or neutrosophic) system, at a given time $\rho$, for each system state $\tau$ one can associate a neutrosophic probability,

$$\mathcal{NP}(\tau) = (t, i, f), \quad (2)$$

where $t, i, f$ are subsets of the unit interval *[0, 1]* such that:

$t$ = the probability that the system resides in $\tau$;

$i$ = the indeterminate probability/improbability about the system residing in $\tau$;

$f$ = the improbability that the system resides in $\tau$;

For a (classical or neutrosophic) dynamic system, the neutrosophic probability of a system's state changes in the time, upon the previous states the system was in, and upon the internal or external conditions.

## 2.36 $(t, i, f)$-reiterative.

In *Neutrosophic Reiterative System*, each state is partially dependent on the previous state. We call this process quasi-reiteration or $(t, i, f)$ *-reiteration.*

In a more general case, each state is partially dependent on the previous *n* states, for $n \geq 1$. This is called *n*-quasi-reiteration, or n-$(t, i, f)$-reiteration.

Therefore, the previous *neutrosophic system history* partialy influences the future neutrosophic system's states, which may be different even if the neutrosophic system started under the same initial conditions.

## 2.37 Finite and infinite system.

A *system* is *finite* if its space, the number of its elements, and the number of its relationships are all finite.

If at least one of these three is infinite, the *system* is considered *infinite*. An infinite system may be countable (if both the number of its elements and the number of its relationships are countable), or, otherwise, uncountable.

## 2.38 Thermodynamic $(t, i, f)$-equilibrium.

The *potential energy* (the work done for changing the system to its present state from its standard configuration) of the classical system is a minimum if the equilibrium is stable, zero if the equilibrium is neutral, or a maximum if the equilibrium is unstable.

A classical system may be in stable, neutral, or unstable equilibrium. A neutrosophic system may be in *quasi-stable*, *quasi-neutral* or *quasi-unstable equilibrium*, and its potential energy respectively quasi-minimum, quasi-null (i.e. close to zero), or quasi-maximum. {We recall that 'quasi' means relative(ly), approximate(ly), almost, near, partial(ly), etc. or mathematically 'quasi' means *(t,i,f)* in a neutrophic way.}

In general, we say that a neutrosophic system is in $(t,i,f)$ – *equilibrium*, or *t%* in stable equilibrium, *i%* in neutral equilibrium, and *f%* in unstable equilibrium (non-equilibrium).

When $f \gg t$ ($f$ is much greater than $t$), the neutroophic system gets into deep non-equilibrium and the perturbations overtake the system's organization to a new organization.

Thus, similarly to the second law of thermodynamics, the neutrosophic system runs down to a $(t,i,f)$-equilibrium state. A neutrosophic system is considered at a thermodynamic $(t,i,f)$-equilibrium state when there is not (or insignificant) flow from a region to another region, and the momentum and energy are uninformally at $(t,i,f)$-level.

## 2.39 The $(t_1, i_1, f_1)$-cause produces a $(t_2, i_2, f_2)$-effect.

In a neutrosophic system, a $(t_1, i_1, f_1)$-cause produces a $(t_2, i_2, f_2)$-effect. We also have cascading $(t, i, f)$-effects from a given cause, and we have permanent change into the system.

$(t, i, f)$-*principles* and $(t, i, f)$-*laws* function in a neutrosophic dynamic system. It is endowed with $(t, i, f)$-*invariants* and with parameters of $(t, i, f)$-*potential* (potentiality, neutrality, impotentiality) control.

## 2.40 $(t, i, f)$-holism.

A neutrosophic system is a $(\boldsymbol{t}, \boldsymbol{i}, \boldsymbol{f})$**-holism**, in the sense that it has a degree of independent entity ($t$) with respect to its parts, a degree of indeterminate ($i$) independent-dependent entity with respect to its parts, and a degree of dependent entity ($f$) with respect to its parts.

## 2.41 Neutrosophic soft assembly.

Only several ways of assembling (combining and arranging) the neutrosophic system's parts are quasi-stable. The others assemble ways are *quasi-transitional*.

The neutrosophic system development is viewed as a *neutrosophic soft assembly*. It is alike an amoeba that changes its shape. In a neutrosophic dynamic system, the space, the elements, the relationships are all flexible, changing, restructuring, reordering, reconnecting and so on, due to heterogeneity, multimodal processes, multi-causalities, multidimensionality, auto-stabilization, auto-hierarchization, auto-embodiement and especially due to synergetism (the neutrosophic system parts cooperating in a $(t, i, f)$-degree).

## 2.42 Neutrosophic collective variable.

The neutrosophic system is partially incoherent (because of the indeterminacy), and partially coherent. Its quasi-behavior is given by the *neutrosophic collective variable* that embeds all neutrosophic variables acting into the $(t, i, f)$-holism.

## 2.43 Conclusion.

We have introduced for the first time notions of neutrosophic system and neutrosophic dynamic system. Of course, these proposals and studies are not exhaustive.

Future investigations have to be done about the neutrosophic (dynamic or not) system, regarding: the neutrosophic descriptive methods and neutrosophic experimental methods, developmental and study the neutrosophic differential equations and neutrosophic difference equations, neutrosophic simulations, the extension of the classical A-Not-B Error to the neutrosophic form, the neutrosophic putative control parameters, neutrosophic loops or neutrosophic cyclic alternations within the system, neutrosophic degenerating (dynamic or not) systems, possible programs within the neutrosophic system, from neutrosophic antecedent conditions how to predict the outcome, also how to find the boundary of neutrosophic conditions, when the neutrosophic invariants are innate/genetic, what are the relationships between the neutrosophic attractors and the neutrosophic repellors, etc.

## 2.44 References.

# 3 Neutrosophic Axiomatic System

## 3.1 Abstract.

In this chapter, we introduce for the first time the notions of *Neutrosophic Axiom, Neutrosophic Deducibility, Neutrosophic Axiomatic System, Neutrosophic Deducibility and Neutrosophic Inference, Neutrosophic Proof, Neutrosophic Tautologies, Neutrosophic Quantifiers, Neutrosophic Propositional Logic, Neutrosophic Axiomatic Space, Degree of Contradiction (Dissimilarity) of Two Neutrosophic Axioms, and Neutrosophic Model*.

A class of neutrosophic implications is also introduced.

A comparison between these innovatory neutrosophic notions and their corresponding classical notions is also made.

Then, three concrete examples of neutrosophic axiomatic systems, describing the same neutrosophic geometrical model, are presented at the end of the chapter.

## 3.2 Neutrosophic Axiom.

A *neutrosophic axiom* or *neutrosophic postulate* (α) is a partial premise, which is *T%* true (degree of truth), *I%* indeterminacy (degree of indeterminacy) and *F%* false (degree of falsehood), where $<t, i, f>$ are standard or nonstandard subsets included in the non-standard unit interval $]^-0, 1^+[$.

The non-standard subsets and non-standard unit interval are mostly used in philosophy in cases where one needs to make distinction between “absolute truth” (which is a truth in all possible worlds) and “relative truth” (which is a truth in at least one world, but not in all possible worlds), and similarly for distinction between “absolute indeterminacy” and “relative indeterminacy”, and respectively distinction between “absolute false-hood” and “relative falsehood”.

But for other scientific and technical applications one uses standard subsets, and the standard classical unit interval *[0, 1]*.

As a particular case of neutrosophic axiom is the classical axiom. In the classical mathematics an axiom is supposed *100%* true, *0%* indeterminate, and *0%* false.

But this thing occurs in idealistic systems, in perfectly closed systems, not in many of the real world situations.

Unlike the classical axiom which is a total premise of reasoning and without any controversy, the neutrosophic axiom is a partial premise of reasoning with a partial controversy.

The neutrosophic axioms serve in approximate reasoning.

The partial truth of a neutrosophic axiom is similarly taken for granted.

The neutrosophic axioms, and in general the neutrosophic propositions, deal with approximate ideas or with probable ideas, and in general with ideas we are not able to measure exactly. That's why one cannot get *100%* true statements (propositions).

In our life, we deal with approximations. An axiom is approximately true, and inference is approximately true either.

A neutrosophic axiom is a self-evident assumption in some degrees of truth, indeterminacy, and falsehood respectively.

## 3.3 Neutrosophic Deducing and Neutrosophic Inference.

The neutrosophic axioms are employed in *neutrosophic deducing* and *neutrosophic inference* rules, which are sort of neutrosophic implications, and similarly they have degrees of truth, indeterminacy, and respectively falsehood.

## 3.4 Neutrosophic Proof.

Consequently, a *neutrosophic proof* has also a degree of validity, degree of indeterminacy, and degree of invalidity. And this is when we work with not-well determinate elements in the space or not not-well determinate inference rules.

The neutrosophic axioms are at the foundation of various *neutrosophic sciences*.

The approximate, indeterminate, incomplete, partially unknown, ambiguous, vagueness, imprecision, contradictory, etc. knowledge can be neutrosophically axiomized.

## 3.5 Neutrosophic Axiomatic System.

A set of neutrosophic axioms $\Omega$, is called *neutrosophic axiomatic system*, where the neutrosophic deducing and the neutrosophic inference (neutrosophic implication) are used.

The neutrosophic axioms are defined on a given space *S*. The space can be classical (space without indeterminacy), or neutrosophic space (space which has some indeterminacy with respect to its elements).

A neutrosophic space may be, for example, a space that has at least one element which only partially belongs to the space. Let us say the element *x<0.5, 0.2, 0.3>* that belongs only *50%* to the space, while *20%* its appurtenance is indeterminate, and *30%* it does not belong to the space.

Therefore, we have three types of neutrosophic axiomatic systems:

1. Neutrosophic axioms defined on classical space;
2. Classical axioms defined on neutrosophic space;
3. Neutrosophic axioms defined on neutrosophic space.

### 3.5.1 Remark.

The neutrosophic axiomatic system is not unique, in the sense that several different axiomatic systems may describe the same neutrosophic model. This happens because one deals with approximations, and because the neutrosophic axioms represent partial (not total) truths.

## 3.6 Classification of the Neutrosophic Axioms.

1) *Neutrosophic Logical Axioms*, which are neutrosophic statements whose truth-value is $<t, i, f>$ within the system of neutrosophic logic. For example: *(α or β)* neutrosophically implies *β*.
2) *Neutrosophic Non-Logical Axioms*, which are neutrosophic properties of the elements of the space. For example: the neutrosophic associativity $a(bc) = (ab)c$, which occurs for some elements, it is unknown (indeterminate) for others, and does not occur for others.

   In general, a neutrosophic non-logical axiom is a classical non-logical axiom that works for certain space elements, is indeterminate for others, and does not work for others.

## 3.7 Neutrosophic Tautologies.

A classical tautology is a statement that is universally true [regarded in a larger way, i.e. lato sensu], i.e. true in all possible worlds (according to Leibniz's definition of "world").

For example "$M = M$" in all possible worlds.

A *neutrosophic tautology* is a statement that is true in a narrow way [i.e. regarded in stricto sensu], or it is $<1, 0, 0>$ true for a class of certain parameters and conditions, and <*t, i, f*> true for another class of certain parameters and conditions, where <*t, i, f*> ≠ <*1, 0, 0*>*.* I.e. a neutrosophic tautology is true in some worlds, and partially true in other worlds. For example the previous assertation: "*M* = *M*".

If "*M*" is a number [i.e. the parameter = number], then a number is always equal to itself in any numeration base.

But if "*M*" is a person [i.e. the parameter = person], call him Martin, then Martin at time $t_1$ is the same as Martin at time $t_1$ [i.e. it has been considered another parameter = time], but Martin at time $t_1$ is different from Martin at time $t_2$ (meaning for example 20 years ago: hence Martin younger is different from Martin older). Therefore, from the point of view of parameters 'person' and 'time', "*M* = *M*" is not a classical tautology.

Similarly, we may have a proposition *P* which is true locally, but it is untrue non-locally.

A neutrosophic logical system is an approximate minimal set of partially true/ indeterminate/ false propositions. While the classical axioms cannot be deduced from other axioms, there are neutrosophic axioms that can be partially deduced from other neutrosophic axioms.

## 3.8 Notations regarding the Classical Logic and Set, Fuzzy Logic and Set, Intuitionistic Fuzzy Logic and Set, and Neutrosophic Logic and Set.

In order to make distinction between classical (Boolean) logic/set, fuzzy logic/set, intuitionistic fuzzy logic/set, and neutrosophic logic/set, we denote their corresponding operators (negation/complement, conjunction/intersection, disjunction/union, implication, and equivalence), as it follows:

a. For classical (Boolean) logic and set:

$$\neg \quad \wedge \quad \vee \quad \rightarrow \quad \leftrightarrow$$

b. For fuzzy logic and set:

$$\underset{F}{\neg} \quad \underset{F}{\wedge} \quad \underset{F}{\vee} \quad \underset{F}{\rightarrow} \quad \underset{F}{\leftrightarrow}$$

c. For intuitionistic fuzzy logic and set:

$$\underset{IF}{\neg} \quad \underset{IF}{\wedge} \quad \underset{IF}{\vee} \quad \underset{IF}{\rightarrow} \quad \underset{IF}{\leftrightarrow}$$

d. For neutrosophic logic and set:

$$\underset{N}{\neg} \quad \underset{N}{\wedge} \quad \underset{N}{\vee} \quad \underset{N}{\rightarrow} \quad \underset{N}{\leftrightarrow}$$

## 3.9 The Classical Quantifiers.

The classical *Existential Quantifier* is the following way:

$$\exists x \in A, P(x). \quad (3)$$

In a neutrosophic way we can write it as:

There exist $x<1, 0, 0>$ in $A$ such that $P(x)<1, 0, 0>$, or

$$\exists x <1,0,0> \in A, P(x) <1,0,0>. \quad (4)$$

The classical *Universal Quantifier* is the following way:

$$\forall x \in A, P(x). \quad (5)$$

In a neutrosophic way we can write it as: For any $x<1, 0, 0>$ in $A$ one has $P(x)<1, 0, 0>$, or

$$\forall x <1,0,0> \in A, P(x) <1,0,0>. \quad (6)$$

## 3.10 The Neutrosophic Quantifiers.

The *Neutrosophic Existential Quantifier* is in the following way:

There exist $x<t_x, i_x, f_x>$ in $A$ such that $P(x)<t_P, i_P, f_P>$, or

$$\exists x < t_x, i_x, f_x > \in A, P(x) < t_P, i_P, f_P >, \quad (7)$$

which means that: there exists an element $x$ which belongs to $A$ in a neutrosophic degree $<t_x, i_x, f_x>$, such that the proposition $P$ has the neutrosophic degree of truth $<t_P, i_P, f_P>$.

The *Neutrosophic Universal Quantifier* is the following way: For any $x<t_x, i_x, f_x>$ in $A$ one has $P(x)<t_P, i_P, f_P>$, or

$$\forall x < t_x, i_x, f_x > \in A, P(x) < t_P, i_P, f_P >, \quad (8)$$

which means that: for any element x that belongs to A in a neutrosophic degree $<t_x, i_x, f_x>$, one has the proposition $P$ with the neutrosophic degree of truth $<t_P, i_P, f_P>$.

## 3.11 Neutrosophic Axiom Schema.

A neutrosophic axiom schema is a neutrosophic rule for generating infinitely many neutrosophic axioms.

Examples of neutrosophic axiom schema:

1) *Neutrosophic Axiom Scheme for Universal Instantiation.*
   Let Φ(x) be a formula, depending on variable x defined on a domain D, in the first-order language L, and let's substitute x for $a \in$ D.

Then the new formula:

$$\forall x \Phi(x) \rightarrow_N \Phi(a) \qquad (9)$$

is $< t_{\rightarrow_N}, i_{\rightarrow_N}, f_{\rightarrow_N} >$ -neutrosophically [universally] valid.

This means the following:

if one knows that a formula Φ(x) holds $<t_x, i_x, f_x>$-neutrosophically for every $x$ in the domain $D$, and for $x = a$ the formula $\Phi(a)$ holds $<t_a, i_a, f_a>$-neutrosophically, then the whole new formula (a) holds $< t_{\rightarrow_N}, i_{\rightarrow_N}, f_{\rightarrow_N} >$ -neutrosophically, where $t_{\rightarrow_N}$ means the truth degree, i $_{\rightarrow_N}$ the indeterminacy degree, and f$_{\rightarrow_N}$ the falsehood degree -- all resulted from the neutrosophic implication$\rightarrow_N$ .

2) *Neutrosophic Axiom Scheme for Existential Generalization.*
   Let Φ*(x)* be a formula, depending on variable *x* defined on a domain *D*, in the first-order language *L*, and let's substitute x for $a \in$ D.

Then the new formula:

$$\Phi(a) \rightarrow_N \exists x \Phi(x) \tag{10}$$

is $< t_{\rightarrow_N}, i_{\rightarrow_N}, f_{\rightarrow_N} >$ -neutrosophically [universally] valid. This means the following: if one knows that a formula Φ*(a)* holds $<t_a, i_a, f_a>$-neutrosophically for a given x = a in the domain *D*, and for every *x* in the domain formula Φ(x) holds $<t_x, i_x, f_x>$-neutrosophically, then the whole new formula (b) holds $< t_{\rightarrow_N}, i_{\rightarrow_N}, f_{\rightarrow_N} >$ -neutrosophically, where t $_{\rightarrow_N}$ means the truth degree, i $_{\rightarrow_N}$ the indeterminacy degree, and f$_{\rightarrow_N}$ the falsehood degree -- all resulted from the neutrosophic implication$\rightarrow_N$ .

These are *neutrosophic metatheorems* of the mathematical neutrosophic theory where they are employed.

## 3.12 Neutrosophic Propositional Logic.

We have many neutrosophic formulas that one takes as neutrosophic axioms. For example, as extension from the classical logic, one has the following.

Let $P<t_P, i_P, f_P>$, $Q<t_Q, i_Q, f_Q>$, $R<t_R, i_R, f_R>$, $S<t_S, i_S, f_S>$ be neutrosophic propositions, where $<t_P, i_P, f_P>$ is the neutrosophic-truth value of $P$, and similarly for $Q$, $R$, and $S$. Then:

a) *Neutrosophic modus ponens* (**neutrosophic implication elimination**):

$$P \rightarrow_N (Q \rightarrow_N P) \quad (11)$$

b) *Neutrosophic modus tollens* (**neutrosophic law of contrapositive**):

$$((P \rightarrow_N Q) \wedge_N \neg_N Q) \rightarrow_N \neg_N P \quad (1)$$

c) Neutrosophic *disjunctive syllogism* **(neutrosophic disjunction elimination):**

$$((P \vee_N Q) \wedge_N \neg_N P) \rightarrow_N Q \quad (2)$$

d) Neutrosophic *hypothetical syllogism* **(neutrosophic chain argument):**

$$((P \rightarrow_N Q) \wedge_N (Q \rightarrow_N R)) \rightarrow_N (P \rightarrow_N R) \quad (3)$$

e) *Neutrosophic constructive dilemma* **(neutrosophic disjunctive version of modus ponens):**

$$(((P \rightarrow_N Q) \wedge_N (R \rightarrow_N S)) \wedge_N (P \vee_N R)) \rightarrow_N (Q \vee_N S)$$
(4)

f) *Neutrosophic destructive dilemma* (**neutrosophic disjunctive version of modus tollens**):

$$(((P \rightarrow_N Q) \wedge_N (R \rightarrow_N S)) \wedge_N (\neg_N Q \vee_N \neg_N S)) \rightarrow_N (\neg_N P \vee_N \neg_N R)$$
(5)

All these neutrosophic formulae also run as *neutrosophic rules of inference*.

These neutrosophic formulas or neutrosophic derivation rules only partially preserve the truth, and depending on the neutrosophic implication operator that is employed the indeterminacy may increase or decrease. This happens for one works with approximations.

While the above classical formulas in classical proportional logic are classical tautologies (i.e. from a neutrosophical point of view they are *100%* true, *0%* indeterminate, and *0%* false), their corresponding neutrosophic formulas are neither classical tautologies nor neutrosophical tautologies, but ordinary neutrosophic propositions whose $< t, i, f >$ – neutrosophic truth-value is resulted from the $\underset{N}{\rightarrow}$ neutrosophic implication

$$A < t_A, i_A, f_A > \underset{N}{\rightarrow} B < (t_B, i_B, f_B) >. \quad (6)$$

### 3.13 Classes of Neutrosophic Negation Operators.

There are defined in neutrosophic literature classes of neutrosophic negation operators as follows: if $A(t_A, i_A, f_A)$, then its negation is:

$\underset{N}{\neg} A(f_A, i_A, t_A)$, or

$\underset{N}{\neg} A(f_A, 1 - i_A, t_A)$, or

$\underset{N}{\neg} A(1 - t_A, 1 - i_A, 1 - f_A)$, or

$\underset{N}{\neg} A(1 - t_A, i_A, 1 - f_A)$ etc. (18)

### 3.14 Classes of Neutrosophic Conjunctive Operators.

Similarly:

if $A(t_A, i_A, f_A)$ and $B(t_B, i_B, f_B)$, then

$A \underset{N}{\wedge} B = \langle t_A \underset{F}{\wedge} t_B, i_A \underset{F}{\vee} i_B, f_A \underset{F}{\vee} f_B \rangle$, (7)

or $A \underset{N}{\wedge} B = \langle t_A \underset{F}{\wedge} t_B, i_A \underset{F}{\wedge} i_B, f_A \underset{F}{\vee} f_B \rangle$, (20)

or $A \underset{N}{\wedge} B = \langle t_A \underset{F}{\wedge} t_B, i_A \underset{F}{\wedge} i_B, f_A \underset{F}{\wedge} f_B \rangle$ (21)

or $A \underset{N}{\wedge} B = \langle t_A \underset{F}{\wedge} t_B, \frac{i_A + i_B}{2}, f_A \underset{F}{\vee} f_B \rangle$, (22)

or $A \underset{N}{\wedge} B = \langle t_A \underset{F}{\wedge} t_B, 1 - \frac{i_A + i_B}{2}, f_A \underset{F}{\vee} f_B \rangle$, (23)

or $A \underset{N}{\wedge} B = \langle t_A \underset{F}{\wedge} t_B, |i_A - i_B|, f_A \underset{F}{\vee} f_B \rangle$, etc. (24)

### 3.15 Classes of Neutrosophic Disjunctive Operators.

And analogously, there were defined:

$A \underset{N}{\vee} B = \langle t_A \underset{F}{\vee} t_B, i_A \underset{F}{\wedge} i_B, f_A \underset{F}{\wedge} f_B \rangle$, (25)

or $A \overset{\vee}{\underset{N}{}} B = \langle t_A \overset{\vee}{\underset{F}{}} t_B, i_A \overset{\vee}{\underset{F}{}} i_B, f_A \overset{\wedge}{\underset{F}{}} f_B \rangle$, (26)

or $A \overset{\vee}{\underset{N}{}} B = \langle t_A \overset{\vee}{\underset{F}{}} t_B, i_A \overset{\vee}{\underset{F}{}} i_B, f_A \overset{\vee}{\underset{F}{}} f_B \rangle$, (27)

or $A \overset{\vee}{\underset{N}{}} B = \langle t_A \overset{\vee}{\underset{F}{}} t_B, \frac{i_A+i_B}{2}, f_A \overset{\wedge}{\underset{F}{}} f_B \rangle$, (28)

or $A \overset{\vee}{\underset{N}{}} B = \langle t_A \overset{\vee}{\underset{F}{}} t_B, 1-\frac{i_A+i_B}{2}, f_A \overset{\wedge}{\underset{F}{}} f_B \rangle$, (29)

or $A \overset{\vee}{\underset{N}{}} B = \langle t_A \overset{\vee}{\underset{F}{}} t_B, |i_A - i_B|, f_A \overset{\vee}{\underset{F}{}} f_B \rangle$, etc. (30)

## 3.16 Fuzzy Operators.

Let $\alpha, \beta \in [0,1]$.

1. The **Fuzzy Negation** has been defined as $\underset{F}{\neg}\alpha = 1 - \alpha$. *(31)*

2. While the class of **Fuzzy Conjunctions** (or **t-norm**) may be:

$\alpha \overset{\wedge}{\underset{F}{}} \beta = \min\{\alpha, \beta\}$, (32)

or $\alpha \overset{\wedge}{\underset{F}{}} \beta = \alpha \cdot \beta$, (33)

or $\alpha \overset{\wedge}{\underset{F}{}} \beta = \max\{0, \alpha + \beta - 1\}$, etc. (34)

3. And the class of **Fuzzy Disjunctions** (or **t-conorm**) may be:

$\alpha \overset{\vee}{\underset{F}{}} \beta = \max\{\alpha, \beta\}$, (35)

or $\alpha \overset{\vee}{\underset{F}{}} \beta = \alpha + \beta - \alpha\beta$, (36)

or $\alpha \overset{\vee}{\underset{F}{}} \beta = \min\{1, \alpha + \beta\}$, etc. (37)

4. Examples of **Fuzzy Implications** $x \underset{F}{\rightarrow} y$, for $x, y \in [0,1]$, defined below:

- Fodor (1993):

$$I_{FD}(x,y) = \begin{cases} 1, & \text{if } x \le y \\ \max(1-x,y), & \text{if } x > y \end{cases} \quad (38)$$

- Weber (1983):

$$I_{WB}(x,y) = \begin{cases} 1, \text{if } x < y \\ y, \text{if } x = 1 \end{cases} \quad (39)$$

- Yager (1980):

$$I_{YG}(x,y) = \begin{cases} 1, \text{if } x = 0 \text{ and } y = 0 \\ y^x, \text{if } x > 0 \text{ or } y > 0 \end{cases} \quad (40)$$

- Goguen (1969):

$$I_{GG}(x,y) = \begin{cases} 1, \text{if } x \le y \\ \frac{y}{x}, \text{if } x > y \end{cases} \quad (41)$$

- Rescher (1969):

$$I_{RS}(x,y) = \begin{cases} 1, \text{if } x \le y \\ 0, \text{if } x > y \end{cases} \quad (42)$$

- Kleene-Dienes (1938):

$$I_{KD}(x,y) = \max(1-x,y) \quad (43)$$

- Reichenbach (1935):

$$I_{RC}(x,y) = 1 - x + xy \quad (44)$$

- Gödel (1932):

$$I_{GD}(x,y) = \begin{cases} 1, \text{if } x \le y \\ y, \text{if } x > y \end{cases} \quad (45)$$

- Lukasiewicz (1923):

$$I_{LK}(x,y) = \min(1, 1-x+y), \quad (46)$$

according to the list made by Michal Baczynski and Balasubramaniam Jayaram (2008).

5. An example of **Intuitionistic Fuzzy Implication** $A(t_A, f_A) \underset{IF}{\rightarrow} B(t_B, f_B)$ is:

$$I_{IF} = \left( [(1-t_A) \underset{F}{\vee} t_B] \underset{F}{\wedge} [(1-f_B) \underset{F}{\vee} f_A], f_B \underset{F}{\wedge} (1-t_A) \right), \qquad (47)$$

according to Yunhua Xiao, Tianyu Xue, Zhan'ao Xue, and Huiru Cheng (2011).

## 3.17 Classes of Neutrosophic Implication Operators.

We now propose for the first time *eight new classes of neutrosophic implications* and extend a ninth one defined previously:

$$A(t_A, i_A, f_A) \underset{N}{\rightarrow} B(t_B, i_B, f_B),$$

in the following ways:

3.17.1-3.17.2 $I_{N1}\left(t_A \underset{F/IF}{\longrightarrow} t_B, i_A \underset{F}{\wedge} i_B, f_A \underset{F}{\wedge} f_B\right)$, (48)

where $t_A \underset{F/IF}{\longrightarrow} t_B$ is any fuzzy implication (from above or others) or any intuitionistic fuzzy implication (from above or others), while $\underset{F}{\wedge}$ is any fuzzy conjunction (from above or others);

3.17.3-3.17.4 $I_{N2}\left(t_A \underset{F/IF}{\longrightarrow} t_B, i_A \underset{F}{\vee} i_B, f_A \underset{F}{\wedge} f_B\right)$, (49)

where $\underset{F}{\vee}$ is any fuzzy disjunction (from above or others);

3.17.5-3.17.6 $I_{N3}\left(t_A \underset{F/IF}{\longrightarrow} t_B, \frac{i_A+i_B}{2}, f_A \underset{F}{\wedge} f_B\right)$; (50)

3.17.7-3.17.8 $I_{N4}\left(t_A \underset{F/IF}{\longrightarrow} t_B, \frac{i_A+i_B}{2}, \frac{f_A+f_B}{2}\right)$. (51)

3.17.9 Now we extend another neutrosophic implication that has been defined by S. Broumi & F. Smarandache (2014) and it was based on the classical logical equivalence:

$$(A \rightarrow B) \leftrightarrow (\neg A \vee B). \quad (52)$$

Whence, since the corresponding neutrosophic logic equivalence:

$$\left(A \underset{N}{\rightarrow} B\right) \underset{N}{\leftrightarrow} (\underset{N}{\neg} A \underset{N}{\vee} B) \quad (53)$$

holds, one obtains another *Class of Neutrosophic Implication Operators* as:

$$(\underset{N}{\neg} A \underset{N}{\vee} B) \quad (54)$$

where one may use any neutrosophic negation $\underset{N}{\neg}$ (from above or others), and any neutrosophic disjunction $\underset{N}{\vee}$ (from above or others).

## 3.18 Example of Neutrosophic Implication.

Let's have two neutrosophic propositions $A\langle 0.3, 0.4, 0.2\rangle$ and $B\langle 0.7, 0.1, 0.4\rangle$. Then $A \underset{N}{\rightarrow} B$ has the neutrosophic truth value of $\underset{N}{\neg} A \underset{N}{\vee} B$, i.e.:

$\langle 0.2, 0.4, 0.3\rangle \underset{N}{\vee} \langle 0.7, 0.1, 0.4\rangle$,

or $\langle \max\{0.2, 0.7\}, \min\{0.4, 0.1\}, \min\{0.3, 0.4\}\rangle$,

or $\langle 0.7, 0.1, 0.3\rangle$,

where we used the neutrosophic operators defined above:

$$\underset{N}{\neg}\langle t, i, f\rangle = \langle f, i, t\rangle$$ for neutrosophic negation

and

$$\langle t_1, i_1, f_1\rangle \underset{N}{\vee} \langle t_2, i_2, f_2\rangle = \langle \max\{t_1, t_2\}, \min\{i_1, i_2\}, \min\{f_1, f_2\}\rangle$$

for the neutrosophic disjunction.

Using different versions of the neutrosophic negation operators and/or different versions of the neutrosophic disjunction operators, one obtains, in general, different results. Similarly as in fuzzy logic.

## 3.19 Another Example of Neutrosophic Implication.

Let $A$ have the neutrosophic truth-value $(t_A, i_A, f_A)$, and $B$ have the neutrosophic truth-value $(t_B, i_B, f_B)$, then:

$$\left[A \underset{N}{\rightarrow} B\right] \underset{N}{\leftrightarrow} [(\underset{N}{\neg} A) \underset{N}{\vee} B], \tag{8}$$

where $\underset{N}{\neg}$ is any of the above neutrosophic negations, while $\underset{N}{\vee}$ is any of the above neutrosophic disjunctions.

## 3.20 General Definition of Neutrosophic Operators.

We consider that the most general definition of neutrosophic operators shall be the followings:

$$A(t_A, i_A, f_A) \underset{N}{\oplus} B(t_B, i_B, f_B) = A \underset{N}{\oplus} B \langle u(t_A, i_A, f_A, t_B, i_B, f_B),$$
$$v(t_A, i_A, f_A, t_B, i_B, f_B),\ w(t_A, i_A, f_A, t_B, i_B, f_B) \rangle \qquad (56)$$

where $\underset{N}{\oplus}$ is any binary neutrosophic operator, and

$$u(x_1, x_2, x_3, x_4, x_5, x_6),\ v(x_1, x_2, x_3, x_4, x_5, x_6),$$
$$w(x_1, x_2, x_3, x_4, x_5, x_6): [0,1]^6 \rightarrow [0,1]. \qquad (57)$$

Even more, the neutrosophic component functions $u, v, w$ may depend, on the top of these six variables, on hidden parameters as well, such as: $h_1, h_2, \dots, h_n$.

For a unary neutrosophic operator (for example, the neutrosophic negation), similarly:

$$\underset{N}{\urcorner} A(t_A, i_A, f_A) =$$
$$\langle u'(t_A, i_A, f_A), v'(t_A, i_A, f_A), w'(t_A, i_A, f_A) \rangle \qquad (58)$$

where

$$u'(t_A, i_A, f_A), v'(t_A, i_A, f_A), w'(t_A, i_A, f_A): [0, 1]^3 \rightarrow [0,1],$$

and even more $u', v', w'$ may depend, on the top of these three variables, of hidden parameters as well, such as: $h_1, h_2, \dots, h_n$.

{Similarly there should be for a *general definition of fuzzy operators* and *general definition of intuitionistic fuzzy operators*.}

As an example, we have defined in F. Smarandache, V. Christianto, *n-ary Fuzzy Logic and Neutrosophic Logic Operators,* published in Studies in

Logic Grammar and Rhetoric, Belarus, 17(30), pp. 1-16, 2009:

$$A(t_A, i_A, f_A)\overset{\wedge}{_N}B(t_B, i_B, f_B) = \langle t_A t_B, i_A i_B + t_A i_B + t_B i_A, t_A f_B + t_B f_A + i_A f_B + i_B f_A \rangle \quad (59)$$

these result from multiplying

$$(t_A + i_A + f_A) \cdot (t_B + i_B + f_B) \quad (60)$$

and ordering upon the below pessimistic order:

truth ≺ indeterminacy ≺ falsity,

meaning that to the *truth* only the terms of $t$'s goes, i.e. $t_A t_B$, to *indeterminacy* only the terms of t's and i's go, i.e. $i_A i_B + t_A i_B + t_B i_A$, and to *falsity* the other terms left, i.e. $t_A f_B + t_B f_A + i_A f_B + i_B f_A + f_A f_B$.

## 3.21 Neutrosophic Deductive System.

A *Neutrosophic Deductive System* consists of a set $\mathcal{L}_1$ of neutrosophic logical axioms, and a set $\mathcal{L}_2$ of neutrosophic non-logical axioms, and a set $\mathcal{R}$ of neutrosophic rules of inference – all defined on a neutrosophic space $\mathcal{S}$ that is composed of many elements.

A neutrosophic deductive system is said to be neutrosophically complete, if for any neutrosophic formula $\varphi$ that is a neutrosophic logical consequence of $\mathcal{L}_1$, i.e. $\mathcal{L}_1 \overset{\vDash}{_N} \varphi$, there exists a neutrosophic deduction of $\varphi$ from $\mathcal{L}_1$, i.e. $\mathcal{L}_1 \overset{\vdash}{_N} \varphi$, where $\underset{N}{\vDash}$ denotes neutrosophic logical consequence, and $\underset{N}{\vdash}$ denotes neutrosophic deduction.

Actually, everything that is neutrosophically (partially) true [i.e. made neutrosophically (partially) true by the set $\mathcal{L}_1$ of neutrosophic axioms] is neutrosophically (partially) provable.

The neutrosophic completeness of set $\mathcal{L}_2$ of neutrosophic non-logical axioms is not the same as the neutrosophic completeness of set $\mathcal{L}_1$ of neutrosophic logical axioms.

## 3.22 Neutrosophic Axiomatic Space.

The space $\mathcal{S}$ is called *neutrosophic space* if it has some indeterminacy with respect to one or more of the following:

a. Its *elements*;
1. At least one element $x$ partially belongs to the set $\mathcal{S}$, or $x(t_x, i_x, f_x) \neq (1, 0, 0)$;
2. There is at least an element $y$ in $\mathcal{S}$ whose appurtenance to $\mathcal{S}$ is unknown.

b. Its *logical axioms*;
1. At least a logical axiom $\mathcal{A}$ is partially true, or $\mathcal{A}(t_A, i_A, f_A)$ , where similary $(t_A, i_A, f_A) \neq (1, 0, 0)$;
2. There is at least an axiom $\mathcal{B}$ whose truth-value is unknown.

c. Its *non-logical axioms*;
1. At least a non-logical axiom $\mathcal{C}$ is true for some elements, and indeterminate or false or other elements;
2. There is at least a non-logical axiom whose truth-value is unknown for some elements in the space.

d. There exist at least two neutrosophic logical axioms that have some degree of contradiction (strictly greater than zero).
e. There exist at least two neutrosophic non-logical axioms that have some degree of contradiction (strictly greater than zero).

## 3.23 Degree of Contradiction (Dissimilarity) of Two Neutrosophic Axioms.

Two neutrosophic logical axioms $\mathcal{A}_1$ and $\mathcal{A}_2$ are contradictory (dissimilar) if their semantics (meanings) are contradictory in some degree $d_1$, while their neutrosophic truth values $<t_1, i_1, f_1>$ and $<t_2, i_2, f_2>$ are contradictory in a different degree $d_2$ [in other words $d_1 \neq d_2$].

As a particular case, if two neutrosophic logical axioms $\mathcal{A}_1$ and $\mathcal{A}_2$ have the same semantic (meaning) [in other words $d_1 = 0$], but their neutrosophic truth-values are different [in other words $d_2 > 0$], they are contradictory.

Another particular case, if two neutrosophic axioms $\mathcal{A}_1$ and $\mathcal{A}_2$ have different semantics (meanings) [in other words $d_1 > 0$], but their neutrosophic truth values are the same $<t_1, i_1, f_1> = <t_2, i_2, f_2>$ [in other words $d_2 = 0$], they are contradictory.

If two neutrosophic axioms $\mathcal{A}_1$ and $\mathcal{A}_2$ have the semantic degree of contradiction $d_1$, and the neutrosophic truth value degree of contradiction $d_2$, then the total degree of contradiction of the two neutrosophic axioms is $d = |d_1 - d_2|$, where $|\ \ |$ mean the absolute value.

We did not manage to design a formula in order to compute the semantic degree of contradiction $d_1$ of two neutrosophic axioms. The reader is invited to explore such metric.

But we can compute the neutrosophic truth value degree of contradiction $d_2$. If $\langle t_1, i_1, f_1 \rangle$ is the neutrosophic truth-value of $\mathcal{A}_1$ and $\langle t_2, i_2, f_2 \rangle$ the neutrosophic truth-value of $\mathcal{A}_2$, where $t_1, i_1, f_1, t_2, i_2, f_2$ are single values in $[0, 1]$, then the neutrosophic truth value degree of contradiction $d_2$ of the neutrosophic axioms $\mathcal{A}_1$ and $\mathcal{A}_2$ is:

$$d_2 = \frac{1}{3}(|t_1 - t_2| + |i_1 - i_2| + |f_1 - f_2|), \quad (61)$$

whence $d_2 \in [0, 1]$.

We get $d_2 = 0$,

when $\mathcal{A}_1$ is identical with $\mathcal{A}_2$ from the point of view of neutrosophical truth values, i.e. when $t_1 = t_2$, $i_1 = i_2$, $f_1 = f_2$.

And we get $d_2 = 1$,

when $\langle t_1, i_1, f_1 \rangle$ and $\langle t_2, i_2, f_2 \rangle$ are respectively equal to:

$\langle 1, 0, 0 \rangle$, $\langle 0, 1, 1 \rangle$;

or $\langle 0, 1, 0 \rangle$, $\langle 1, 0, 1 \rangle$;

or $\langle 0, 0, 1 \rangle$, $\langle 1, 1, 0 \rangle$;

or $\langle 0, 0, 0 \rangle$, $\langle 1, 1, 1 \rangle$.

## 3.24 Neutrosophic Axiomatic System.

The **neutrosophic axioms** are used, in neutrosophic conjunction, in order to derive neutrosophic theorems.

A *neutrosophic mathematical theory* may consist of a neutrosophic space where a neutrosophic axiomatic system acts and produces all neutrosophic theorems within the theory.

Yet, in a *neutrosophic formal system*, in general, the more recurrences are done the more is increased the indeterminacy and decreased the accuracy.

## 3.25 Properties of the Neutrosophic Axiomatic System.

1. While in classical mathematics an axiomatic system is consistent, in a neutrosophic axiomatic system it happens to have *partially inconsistent (contradictory) axioms*.
2. Similarly, while in classical mathematics the axioms are independent, in a neutrosophic axiomatic system they may be *dependent in certain degree*.

   In classical mathematics if an axiom is dependent from other axioms, it can be removed, without affecting the axiomatic system. However, if a neutrosophic axiom is partially dependent from other neutrosophic axioms, by removing it the neutrosophic axiomatic system is affected.
3. While, again, in classical mathematics an axiomatic system has to be complete (meaning that each statement or its negation is derivable), a neutrosophic axiomatic system is *partially complete and partially incomplete*. It is partially incomplete because one can add extra partially independent neutrosophic axioms.

4. The *neutrosophic relative consistency* of an axiomatic system is referred to the neutrosophically (partially) undefined terms of a first neutrosophic axiomatic system that are assigned neutrosophic definitions from another neutrosophic axiomatic system in a way that, with respect to both neutrosophic axiomatic systems, is neutrosophically consistent.

## 3.26 Neutrosophic Model.

A *Neutrosophic Model* is a model that assigns neutrosophic meaning to the neutrosophically (un)defined terms of a neutrosophic axiomatic system.

Similarly to the classical model, we have the following classification:

1. *Neutrosophic Abstract Model*, which is a neutrosophic model based on another neutrosophic axiomatic system.
2. *Neutrosophic Concrete Model*, which is a neutrosophic model based on real world, i.e. using real objects and real relations between the objects.

In general, a neutrosophic model is a *<t, i, f>-approximation*, i.e. *T%* of accuracy, *I%* indeter-minacy, and *F%* inaccuracy, of a neutrosophic axiomatic system.

### 3.27 Neutrosophically Isomorphic Models.

Further, *two neutrosophic models are neutrosophically isomorphic* if there is a neutrosophic one-to-one correspondence between their neutrosophic elements such that their neutrosophic relationships hold.

A neutrosophic axiomatic system is called *neutrosophically categorial* (or *categorical*) is any two of its neutrosophic models are neutrosophically isomorphic.

### 3.28 Neutrosophic Infinite Regressions.

There may be situations of neutrosophic axiomatic systems generating neutrosophic infinite regressions, unlike the classical axiomatic systems.

### 3.29 Neutrosophic Axiomatization.

A *Neutrosophic Axiomatization* is referred to an approximate formulation of a set of neutrosophic statements, about a number of neutrosophic primitive terms, such that by the neutrosophic deduction one obtains various neutrosophic propositions (theorems).

## 3.30 Example of Neutrosophic Axiomatic System.

Let's consider two neighboring countries *M* and *N* that have a disputed frontier zone *Z*:

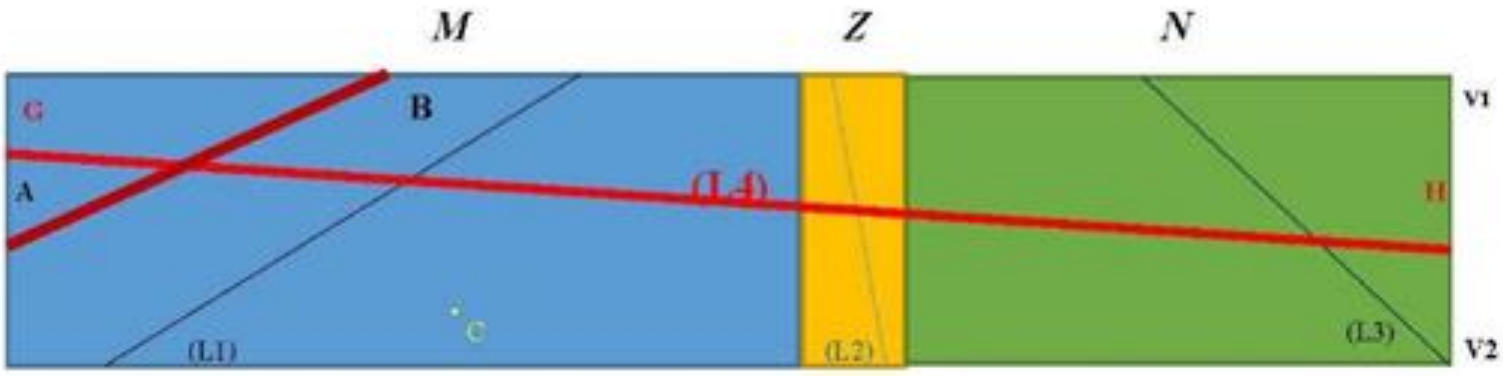

*Fig. 4 A Neutrosophic Model.*

Let's consider the universe of discourse $U = M \cup Z \cup N$; this is a *neutrosophic space* since it has an indeterminate part (the disputed frontier).

The *neutrosophic primitive notions* in this example are: neutrosophic point, neutrosophic line, and neutrosophic plane (space).

And the *neutrosophic primitive relations* are: neutrosophic incidence, and neutrosophic parallel.

The four boundary edges of rectangle *Z* belong to *Z* (or *Z* is a closed set). While only three boundary edges of *M* (except the fourth one which is common with *Z*) belong to *M*, and similarly only three boundaries of *N* (except the fourth one which is common with *Z*) belong to *N*. Therefore *M* and *N* are neither closed nor open sets.

Taking a classical point *P* in *U*, one has three possibilities:

[1] $P \in M$ (membership with respect to country *M*);

[2] $P \in Z$ (indeterminate membership with respect to both countries);

[3] or $P \in N$ (nonmembership with respect to country *M*).

Such points, that can be indeterminate as well, are called *neutrosophic points*.

A *neutrosophic line* is a classical segment of line that unites two neutrosophic points lying on opposite edges of the universe of discourse *U*. We may have:

[1] determinate line (with respect to country *M*), that is completely into the determinate part *M* {for example (*L1*)};

[2] indeterminate line, that is completely into the frontier zone {for example (*L2*)};

[3] determinate line (with respect to country *N*), that is completely into the determinate part *N* {for example (*L3*)};

[4] or mixed, i.e. either two or three of the following: partially determinate

with respect to *M*, partially indeterminate with respect to both countries, and partially determinate with respect to *N {for example the red line (L4)}.*

Through two neutrosophic points there may be passing:

[1] only one neutrosophic line {for example, through *G* and *H* passes only one neutrosophic line (*L4*)};

[2] no neutrosophic line {for example, through *A* and *B* passes no neutrosophic line, since the classical segment of line *AB* does not unite points of opposite edges of the universe of discourse *U*}.

Two *neutrosophic lines* are *parallel* is they have no common neutrosophic points.

Through a neutrosophic point outside of a neutrosophic line, one can draw:

[1] infinitely many neutrosophic parallels {for example, through the neutrosophic point *C* one can draw infinitely many neutrosophic parallels to the neutrosophic line *(L1)*};

[2] only one neutrosophic parallel {for example, through the neutrosophic point *H* that belongs to the edge *(V1V2)* one can draw only one neutrosophic parallel (i.e. V1V2) to the neutrosophic line *(L1)*};

[3] no neutrosophic parallel {for example, through the neutrosophic point *H* there is no neutrosophic parallel to the neutrosophic line *(L3)*}.

For example, the neutrosophic lines (*L1), (L2)* and (*L3)* are parallel. But the neutrosophic line *(L4)* is not parallel with *(L1),* nor with *(L2)* or *(L3)*.

A *neutrosophic polygon* is a classical polygon which has one or more of the following indeterminacies:

[1] indeterminate vertex;

[2] partially or totally indeterminate edge;

[3] partially or totally indeterminate region in the interior of the polygon.

We may construct several neutrosophic axiomatic systems, for this example, referring to incidence and parallel.

a) <u>First neutrosophic axiomatic system</u>.

*α1)* Through two distinct neutrosophic points there is passing a single neutrosophic line.

{According to several experts, the neutrosophic truth-value of this axiom is *<0.6, 0.1, 0.2>,* meaning that having two given neutrosophic points, the chance that only one line (that do not intersect the indeterminate zone *Z*) passes through them is *0.6,* the chance that line that passes through them intersects the indeterminate zone *Z*) is *0.1*, and the chance that no line (that does not intersect the indeterminate zone *Z*) passes through them is *0.2*.}

*α2)* Through a neutrosophic point exterior to a neutrosophic line there is passing either one neutrosophic parallel or infinitely many neutrosophic parallels.

{According to several experts, the neutrosophic truth-value of this axiom is *<0.7, 0.2, 0.3>,* meaning that having a given neutrosophic line and a given exterior neutrosophic point, the chance that infinitely many parallels pass through this exterior point is *0.7*, the chance that the parallels passing through this exterior point intersect the indeterminate zone *Z* is

*0.2*, and the chance that no parallel passes through this point is *0.3.*}

Now, let's apply a first neutrosophic deducibility.

Suppose one has three non-collinear neutrosophic (distinct) points *P, Q,* and *R* (meaning points not on the same line, alike in classical geometry). According to the neutrosophic axiom *(α1),* through *P, Q* passes only one neutrosophic line {let's call it *(PQ)*}, with a neutrosophic truth value *(0.6, 0.1, 0.2).* Now, according to axiom *(α2)*, through the neutrosophic point *R,* which does not lie on *(PQ),* there is passing either only one neutrosophic parallel or infinitely many neutrosophic parallels to the neutrosophic line (PQ), with a neutrosophic truth value *(0.7, 0.2, 0.3).*

Therefore,

$$(\alpha 1) \underset{N}{\wedge} (\alpha 2) = <0.6, 0.1, 0.2> \underset{N}{\wedge} <0.7, 0.2, 0.3> = <\min\{0.6, 0.7\}, \max\{0.1, 0.2\}, \max\{0.2, 0.3\}> = <0.6, 0.2, 0.3>,$$

which means the following: the chance that through the two distinct given neutrosophic points *P* and *Q* passes only one neutrosophic line, and through the exterior neutrosophic point *R* passese either only one neutrosophic parallel or infinitely many parallels to *(PQ)* is *(0.6, 0.2, 0.3),* i.e. *60%* true, *20%* indeterminate, and *30%* false.

Herein we have used the simplest neutrosophic conjunction operator $\underset{N}{\wedge}$ of the form *<min, max, max>*, but other neutrosophic conjunction operator can be used as well.

A second neutrosophic deducibility:

Again, suppose one has three non-collinear neutrosophic (distinct) points *P, Q,* and *R* (meaning points not on the same line, as in classical geometry).

Now, let's compute the neutrosophic truth value that through *P* and *Q* is passing one neutrosophic line, but through *Q* there is no neutrosophic parallel to *(PQ)*.

$$\alpha 1 \underset{N}{\wedge} (\underset{N}{\neg} \alpha 2) = <0.6, 0.1, 0.2> \underset{N}{\wedge} (\underset{N}{\neg} <0.7, 0.2, 0.3>) = <0.6, 0.1, 0.2> \underset{N}{\wedge} <0.3, 0.2, 0.7> = <0.3, 0.2, 0.7>.$$

b) Second neutrosophic axiomatic system:

*β1)* Through two distinct neutrosophic points there is passing either a single neutrosophic line or no neutrosophic line. {With the neutrosophic truth-value *<0.8, 0.1, 0.0>*}.

*β2)* Through a neutrosophic point exterior to a neutrosophic line there is passing either one neutrosophic parallel, or infinitely many neutrosophic parallels, or no neutrosophic parallel. {With the neutrosophic truth-value *<1.0, 0.2, 0.0>*}.

In this neutrosophic axiomatic system the above propositions W1 and W2:

> W1: Through two given neutrosophic points there is passing only one neutrosophic line, and through a neutrosophic point exterior to this neutrosophic line there is passing either one neutrosophic parallel or infinitely many neutrosophic parallels to the given neutrosophic line; and W2: Through two given neutrosophic points there is passing only one neutrosophic line, and through a neutrosophic point exterior to this neutrosophic line there is passing no neutrosophic parallel to the line; are not deducible.

c) Third neutrosophic axiomatic system.

*γ1)* Through two distinct neutrosophic points there is passing a single neutrosophic line.
{With the neutrosophic truth-value *<0.6, 0.1, 0.2>*}.

*γ2)* Through two distinct neutrosophic points there is passing no neutrosophic line.
{With the neutrosophic truth-value *<0.2, 0.1, 0.6>*}.

*δ1)* Through a neutrosophic point exterior to a neutrosophic line there is passing only one neutrosophic parallel.
{With the neutrosophic truth-value *<0.1, 0.2, 0.9>*}.

*δ2)* Through a neutrosophic point exterior to a neutrosophic line there are passing infinitely many neutrosophic parallels.
{With the neutrosophic truth-value *<0.6, 0.2, 0.4>*}.

*δ3)* Through a neutrosophic point exterior to a neutrosophic line there is passing no neutrosophic parallel.
{With the neutrosophic truth-value *<0.3, 0.2, 0.7>*}.

In this neutrosophic axiomatic system we have contradictory axioms:

- *(γ1)* is in *100%* degree of contradiction with *(γ2);*
- and similarly *(δ3)* is in *100%* degree of contradiction with *[(δ1)* together with *(δ2)].*

Totally or partially contradictory axioms are allowed in a neutrosophic axiomatic systems, since they are part of our imperfect world and since they approximately describe models that are - in general - partially true.

Regarding the previous two neutrosophic deducibilities one has:

$$\gamma 1 \underset{N}{\wedge} (\delta 1 \underset{N}{\vee} \delta 2) = <0.6, 0.1, 0.2> \underset{N}{\wedge} \left( < 0.1, 0.2, 0.9 > \underset{N}{\vee} < 0.6, 0.2, 0.4 > \right)$$

$$= < 0.6, 0.1, 0.2 > \underset{N}{\wedge} < max\{0.1, 0.6\}, min\{0.2, 0.2\}, min\{0.9, 0.4\} >= < 0.6, 0.1, 0.2 > \underset{N}{\wedge} < 0.6, 0.2, 0.4 >= <0.6, 0.2, 0.4>,$$

which is slightly different from the result we got using the first neutrosophic axiomatic system *<0.6, 0.2, 0.3>*, and respectively*:*

$$\gamma 1 \underset{N}{\wedge} \delta 3 = <0.6, 0.1, 0.2> \underset{N}{\wedge} < 0.3, 0.2, 0.7 >=<0.3, 0.2, 0.7>,$$

which is the same as the result we got using the first neutrosophic axiomatic system.

The third neutrosophic axiomatic system is a refinement of the first and second neutrosophic axiomatic systems. From a deducibility point of view it is better and easier to work with a refined system than with a rough system.

### 3.31 Conclusion.

In many real world situations the spaces and laws are not exact, not perfect. They are inter-dependent. This means that in most cases they are not 100% true, i.e. not universal. For example many physical laws are valid in ideal and perfectly closed systems. But perfectly closed systems do not exist in our heterogeneous world where we mostly deal with approximations. Also, since in the real world there is not a single homogenous space, we have to use the multispace for any attempt to unify various theories.

We do not have perfect spaces and perfect systems in reality. Therefore many physical laws function approximatively (see [5]). The physical constants are not universal too; variations of their values depend from a space to another, from a system to another. A physical constant is t% true, i% indeterminate, and f% false in a given space with a certain composition, and it has a different neutrosophical truth value <t', i', f'> in another space with another composition.

A neutrosophic axiomatic system may be dynamic: new axioms can be added and others excluded. The neutrosophic axiomatic systems are formed by axioms than can be partially dependent (redundant), partially contradictory (inconsistent), partially incomplete, and reflecting a partial truth (and consequently a partial indeterminacy and a partial falsehood) - since they deal with approximations of reality.

## 3.32 References.

# 4 *(t, i, f)*-Neutrosophic Structures & *I*-Neutrosophic Structures

## 4.1 Abstract.

This chapter is an improvement of our paper "(t, i, f)-Neutrosophic Structures" *[1],* where we introduced for the first time a new type of structures, called *(t, i, f)-Neutrosophic Structures*, presented from a neutrosophic logic perspective, and we showed particular cases of such structures in geometry and in algebra.

In any field of knowledge, each structure is composed from two parts: a **space**, and a **set of axioms** (or **laws**) acting (governing) on it. If the space, or at least one of its axioms (laws), has some indeterminacy of the form *(t, i, f)* $\neq$ *(1, 0, 0),* that structure is a *(t, i, f)-Neutrosophic Structure*.

The *(t, i, f)-Neutrosophic Structures* [based on the components *t* = truth, *i* = numerical indeterminacy, *f* = falsehood] are different from the *Neutrosophic Algebraic Structures* [based on neutrosophic numbers of the form $a + bI$, where I = literal indeterminacy and $I^n = I$], that we rename as *I*-

*Neutrosophic Algebraic Structures* (meaning algebraic structures based on indeterminacy "I" only). But we can combine both and obtain the *(t, i, f)-I-Neutrosophic Algebraic Structures*, i.e. algebraic structures based on neutrosophic numbers of the form *a+bI*, but also having indeterminacy of the form *(t, i, f)* ≠ *(1, 0, 0)* related to the structure space (elements which only partially belong to the space, or elements we know nothing if they belong to the space or not) or indeterminacy of the form *(t, i, f)* ≠ *(1, 0, 0)* related to at least one axiom (or law) acting on the structure space. Then we extend them to *Refined (t, i, f)- Refined I-Neutrosophic Algebraic Structures.*

## 4.2 Classification of Indeterminacies.

1. *Numerical Indeterminacy* (or *Degree of Indeterminacy)*, which has the form *(t, i, f)* ≠ *(1, 0, 0)*, where *t, i, f* are numbers, intervals, or subsets included in the unit interval *[0, 1]*, and it is the base for the *(t, i, f)*-Neutrosophic Structures.
2. *Non-numerical Indeterminacy (or Literal Indeterminacy)*, which is the letter "*I*" standing for unknown (non-determinate), such that $I^2 = I$, and used in the composition of the neutrosophic number $N = a + bI$, where *a* and *b* are real or complex numbers, and *a* is the determinate part of number *N*, while *bI* is the indeterminate part of *N*. The neutrosophic numbers are the base for the *I*-Neutrosophic Structures.

## 4.3 Neutrosophic Algebraic Structures [or *I*-Neutrosophic Algebraic Structures].

A previous type of neutrosophic structures was introduced in algebra by W. B. Vasantha Kandasamy and Florentin Smarandache [2-57], since *2003*, and it was called Neutrosophic Algebraic Structures. Later on, more researchers joined the neutrosophic research, such as: Mumtaz Ali, Said Broumi, Jun Ye, A. A. Salama, Muhammad Shabir, K. Ilanthenral, Meena Kandasamy, H. Wang, Y.-Q. Zhang, R. Sunderraman, Andrew Schumann, Salah Osman, D. Rabounski, V. Christianto, Jiang Zhengjie, Tudor Paroiu, Stefan Vladutescu, Mirela Teodorescu, Daniela Gifu, Alina Tenescu, Fu Yuhua, Francisco Gallego Lupiañez, etc.

The neutrosophic algebraic structures are algebraic structures based on sets of **neutrosophic numbers** of the form $N = a + bI$, where *a, b* are real (or complex) numbers, and *a* is called the <u>determinate part</u> on *N* and *bI* is called the indeterminate part of $N$, with $mI + nI = (m + n)I$, $0 \cdot I = 0$, $I^n = I$ for integer $n \geq 1$, and $I / I$ = undefined.

When $a, b$ are real numbers, then $a + bI$ is called a *neutrosophic real number*. While if at least one of $a, b$ is a complex number, then $a + bI$ is called a *neutrosophic complex number*.

We may say "literal indeterminacy" for "$I$" from $a+bI$, and "numerical indeterminacy" for "$i$" from $(t, i, f)$ in order to distinguish them.

The neutrosophic algebraic structures studied by Vasantha-Smarandache in the period *2003-2015* are: neutrosophic groupoid, neutrosophic semigroup, neutrosophic group, neutrosophic ring, neutrosophic field, neutrosophic vector space, neutrosophic linear algebras etc., which later (between *2006-2011*) were generalized by the same researchers to *neutrosophic bi-algebraic structures*, and more general to *neutrosophic N-algebraic structures*.

Afterwards, the neutrosophic structures were further extended to *neutrosophic soft algebraic structures* by Florentin Smarandache, Mumtaz Ali, Muhammad Shabir, and Munazza Naz in 2013-2014.

In 2015 Smarandache refined the literal indeterminacy $I$ into different types of literal indeterminacies (depending on the problem to solve) such as $I_1, I_2, \ldots, I_p$ with integer $p \geq 1$, and obtained the ***refined neutrosophic numbers*** of the form $N_p = a+b_1I_1+b_2I_2+\ldots+b_pI_p$ where $a, b_1, b_2, \ldots, b_p$ are real or complex numbers, and $a$ is called the determinate part of $N_p$, while for each $k \in \{1, 2, \ldots, p\}$ $b_kI_k$ is called the *k*-th indeterminate part of $N_p$, and for each $k \in \{1, 2, \ldots, p\}$, one similarly has: $mI_k + nI_k = (m + n)I_k$, $0 \cdot I_k = 0$, $I_k^n = I_k$ for integer $n \geq 1$, and $I_k/I_k$ = undefined.

The relationships and operations between $I_j$ and $I_k$, for $j \neq k$, depend on each particular problem we need to solve.

Then consequently, Smarandache [2015] extended the neutrosophic algebraic structures to **Refined Neutrosophic Algebraic Structures** [or **Refined *I*-Neutrosophic Algebraic Structures**], which are algebraic structures based on the sets of the refined neutrosophic numbers $a+b_1I_1+b_2I_2+\ldots+b_pI_p$.

## 4.4 *(t, i, f)*-Neutrosophic Structures.

We now introduce for the first time another type of neutrosophic structures. These structures, in any field of knowledge, are considered from a neutrosophic logic point of view, i.e. from the truth-indeterminacy-falsehood $(t, i, f)$ values.

In neutrosophic logic every proposition has a degree of truth *(t)*, a degree of indeterminacy *(i)*, and a degree of falsehood *(f)*, where *t, i, f* are standard or non-standard subsets of the non-standard unit interval $]^{-}0, 1^{+}[$. In technical applications *t, i,* and *f* are only standard subsets of the standard unit interval *[0, 1]* with: $^{-}0 \leq sup(T) + sup(I) + sup(F) \leq 3^{+}$, where *sup(X)* means supremum of the subset *X*.

In general, each structure is composed from: a **space**, endowed with a **set of axioms** (or **laws**) acting (governing) on it. If the space, or at least one of its axioms, has some numerical indeterminacy of the form $(t, i, f) \neq (1, 0, 0)$, we consider it as a ***(t, i, f)*-Neutrosophic Structure**.

Indeterminacy with respect to the space is referred to some elements that partially belong [i.e. with a neutrosophic value $(t, i, f) \neq (1, 0, 0)$] to the space, or their appurtenance to the space is unknown. An axiom (or law) which deals with numerical indeterminacy is called *neutrosophic axiom* (or *law)*. We introduce these new structures because in the real world we do not always know exactly or completely the space we work in; and because the axioms (or laws) are not always well defined on this space, or may have indeterminacies when applying them.

## 4.5 Refined *(t, i, f)*-Neutrosophic Structures [or $(t_j, i_k, f_l)$-Neutrosophic Structures]

In 2013 Smarandache [76] refined the numerical neutrosophic components *(t, i, f)* into $(t_1, t_2, \ldots, t_m;\ i_1, i_2, \ldots, i_p;\ f_1, f_2, \ldots, f_r)$, where $m, p, r$ are integers $\geq 1$.

Consequently, we now [2015] extend the *(t, i, f)*-Neutrosophic Structures to $(t_1, t_2, \ldots, t_m;\ i_1, i_2, \ldots, i_p;\ f_1, f_2, \ldots, f_r)$-Neutrosophic Structures, that we called *Refined (t, i, f)-Neutrosophic Structures* [or $(t_j, i_k, f_l)$-*Neutrosophic Structures*]. These are structures whose elements have a refined neutrosophic value of the form $(t_1, t_2, \ldots, t_m;\ i_1, i_2, \ldots, i_p;\ f_1, f_2, \ldots, f_r)$ or the space has some indeterminacy of this form.

## 4.6 *(t, i, f)-I*-Neutrosophic Algebraic Structures.

The *(t, i, f)-Neutrosophic Structures* [based on the numerical components $t$ = truth, $i$ = indeterminacy, $f$ = falsehood] are different from the *Neutrosophic Algebraic Structures* [based on neutrosophic numbers of the form $a + bI$]. We may rename the last ones as *I-Neutrosophic Algebraic Structures* (meaning: algebraic structures based on literal indeterminacy "*I*" only).

But we can combine both of them and obtain a *(t, i, f)-I-Neutrosophic Algebraic Structures*,

i.e. algebraic structures based on neutrosophic numbers of the form $a + bI$, but this structure also having indeterminacy of the form $(t, i, f) \neq (1, 0, 0)$ related to the structure space (elements which only partially belong to the space, or elements we know nothing if they belong to the space or not) or indeterminacy related to at least an axiom (or law) acting on the structure space. Even more, we can generalize them to **Refined *(t, i, f)*- *Refined I*-Neutrosophic Algebraic Structures**, or **$(t_j, i_k, f_l)$-$I_s$-Neutrosophic Algebraic Structures.**

## 4.7 Example of Refined *I*-Neutrosophic Algebraic Structure.

Let the indeterminacy $I$ be split into $I_1$ = contradiction (i.e. truth and falsehood simultaneously), $I_2$ = ignorance (i.e. truth or falsehood), and $I_3$ = vagueness, and the corresponding *3*-refined neutrosophic numbers of the form $a+b_1I_1+b_2I_2+b_3I_3$.

Let $(G, *)$ be a groupoid. Then the *3-refined I-neutrosophic groupoid* is generated by $I_1, I_2, I_3$ and $G$ under * and it is denoted by

$$N_3(G) = \{(G \cup I_1 \cup I_2 \cup I_3), *\} = \{ a+b_1I_1+b_2I_2+b_3I_3 \;/\; a, b_1, b_2, b_3 \in G \}. \quad (62)$$

## 4.8 Example of Refined *(t, i, f)*-Neutrosophic Structure.

Let *(t, i, f)* be split as $(t_1, t_2; i_1, i_2; f_1, f_2, f_3)$. Let $H = ( \{h_1, h_2, h_3\}, \# )$ be a groupoid, where $h_1$, $h_2$, and $h_3$ are real numbers. Since the elements $h_1$, $h_2$, $h_3$ only partially belong to H in a refined way, we define a *refined (t, i, f)-neutrosophic groupoid* { or *refined (2; 2; 3)-neutrosophic groupoid,* since *t* was split into *2* parts, *I* into *2* parts, and *t* into *3* parts } as

$H = \{h_1(0.1, 0.1; \ 0.3, 0.0; \ 0.2, 0.4, 0.1), h_2(0.0, 0.1; \ 0.2, 0.1; \ 0.2, 0.0, 0.1),$
$h_3(0.1, 0.0; \ 0.3, 0.2; \ 0.1, 0.4, 0.0)\}$.

## 4.9 Examples of *(t, i, f)-I*-Neutrosophic Algebraic Structures.

1) *Indeterminate Space (due to Unknown Element); with Neutrosophic Number included.*

   Let $B = \{2+5I, \ -I, \ -4, \ b(0, \ 0.9, \ 0)\}$ a neutrosophic set, which contains two neutrosophic numbers, *2+5I* and *-I*, and we know about the element *b* that its appurtenance to the neutrosophic set is *90%* indeterminate.

2) *Indeterminate Space (due to Partially Known Element); with Neutrosophic Number included.*

   Let $C = \{-7, 0, 2+I(0.5, 0.4, 0.1), 11(0.9, 0, 0)\}$, which contains a neutrosophic number $2+I$, and this neutrosophic number is actually only partially in $C$; the element $11$ is also partially in $C$.

3) *Indeterminacy Axiom (Law).*

   Let $D = [0+0I, 1+1I] = \{c+dI, \text{ where } c, d \in [0, 1]\}$. One defines the binary law # in the following way:

   $$\# : D \times D \rightarrow D, \; x \# y = (x_1 + x_2 I) \# (y_1 + y_2 I) = [(x_1 + x_2)/y_1] + y_2 I, \qquad (63)$$

   but this neutrosophic law is undefined (indeterminate) when $y_1 = 0$.

4) *Little Known or Completely Unknown Axiom (Law).*

   Let us reconsider the same neutrosophic set $D$ as above. But, about the binary neutrosophic law $\Theta$ that $D$ is endowed with, we only know that it associates the neutrosophic numbers $1+I$ and $0.2+0.3I$ with the neutrosophic number $0.5+0.4I$, i.e. $(1+I)\Theta(0.2+0.3I) = 0.5+0.4I$.

There are many cases in our world when we barely know some axioms (laws).

## 4.10 Examples of Refined *(t, i, f)- Refined I-* Neutrosophic Algebraic Structures.

We combine the ideas from Examples 5 and 6 and we construct the following example.

Let's consider, from Example *5*, the groupoid *(G, *)*, where *G* is a subset of positive real numbers, and its extension to a *3-refined I-neutrosophic groupoid*, which was generated by $I_1$, $I_2$, $I_3$ and *G* under the law * that was denoted by

$$N_3(G) = \{ a+b_1I_1+b_2I_2+b_3I_3 \,/\, a, b_1, b_2, b_3 \in G \}. \quad (64)$$

We then endow each element from $N_3(G)$ with some *(2; 2; 3)*-refined degrees of membership/ indeterminacy/nonmembership, as in Example 6, of the form $(T_1, T_2; I_1, I_2; F_1, F_2, F_3)$, and we obtain a

$$N_3(G)_{(2;2;3)} = \{ a+b_1I_1+b_2I_2+b_3I_3(T_1, T_2; I_1, I_2; F_1, F_2, F_3) \,/\, a, b_1, b_2, b_3 \in G \}, \quad (65)$$

where (66)

$$T_1 = \frac{a}{a+b_1+b_2+b_3}, T_2 = \frac{0.5a}{a+b_1+b_2+b_3};$$

$$I_1 = \frac{b_1}{a+b_1+b_2+b_3}, I_2 = \frac{b_2}{a+b_1+b_2+b_3};$$

$$F_1 = \frac{0.1b_3}{a+b_1+b_2+b_3}, F_2 = \frac{0.2b_1}{a+b_1+b_2+b_3}, F3 = \frac{b_2+b_3}{a+b_1+b_2+b_3}.$$

Therefore, $N_3(G)_{(2;2;3)}$ is a *refined (2; 2; 3)-neutrosophic groupoid* and a *3-refined I-neutrosophic groupoid*.

## 4.11 Neutrosophic Geometric Examples.

a) *Indeterminate Space*.

We might not know if a point *P* belongs or not to a space *S* [we write *P(0, 1, 0),* meaning that *P*'s indeterminacy is *1*, or completely unknown, with respect to *S*].

Or we might know that a point *Q* only partially belongs to the space *S* and partially does not belong to the space *S* [for example *Q(0.3, 0.4, 0.5),* which means that with respect to *S*, *Q*'s membership is *0.3, Q*'s indeterminacy is *0.4*, and *Q*'s non-membership is *0.5*].

Such situations occur when the space has vague or unknown frontiers, or the space contains ambiguous (not well-defined) regions.

b) *Indeterminate Axiom*.

Also, an axiom *(α)* might not be well defined on the space *S*, i.e. for some elements of the space the axiom *(α)* may be valid, for other elements of the space the axiom *(α)* may be indeterminate (meaning neither valid, nor invalid), while for the remaining elements the axiom *(α)* may be invalid.

As a concrete example, let's say that the neutrosophic values of the axiom *(α)* are *(0.6, 0.1, 0.2)* = (degree of validity, degree of indeterminacy, degree of invalidity).

## 4.12 *(t, i, f)*-Neutrosophic Geometry as a Particular Case of *(t, i, f)*-Neutrosophic Structures.

As a particular case of *(t, i, f)*-neutrosophic structures in geometry, one considers a ***(t, i, f)*-Neutrosophic Geometry** as a geometry which is defined either on a space with some indeterminacy (i.e. a portion of the space is not known, or is vague, confused, unclear, imprecise), or at least one of its axioms has some indeterminacy of the form $(t, i, f) \neq (1, 0, 0)$ (i.e. one does not know if the axiom is verified or not in the given space, or for some elements the axiom is verified and for others it is not verified).

This is a generalization of the *Smarandache Geometry* (SG) [57-75], where an axiom is validated and invalidated in the same space, or only invalidated, but in multiple ways. Yet the SG has no degree of indeterminacy related to the space or related to the axiom.

A simple **Example of a SG** is the following – that unites Euclidean, Lobachevsky-Bolyai-Gauss, and Riemannian geometries altogether, in the same space, considering the Fifth Postulate of Euclid: in one region of the SG space the postulate is validated (only one parallel trough a point to a given line), in a second region of SG the postulate is invalidated (no parallel through a point to a given line – elliptical geometry), and in a third region of SG the postulate is invalidated but in a different way (many parallels through a point to a given line – hyperbolic geometry). This simple example shows a hybrid geometry which is partially Euclidean, partially Non-Euclidean Elliptic, and partially Non-Euclidean Hyperbolic. Therefore, the fifth postulate (axiom) of Euclid is true for some regions, and false for others, but it is not indeterminate for any region (i.e. not knowing how many parallels can be drawn through a point to a given line).

We can extend this hybrid geometry adding a new space region where one does not know if there are or there are not parallels through some given points to the given lines (i.e. the Indeterminate component) and we form a more complex *(t, i, f)-Neutrosophic Geometry.*

## 4.13 Neutrosophic Algebraic Examples.

1) *Indeterminate Space*
*(due to Unknown Element).*

Let the set (space) be $NH = \{4, 6, 7, 9, a\}$, where the set $NH$ has an unknown element "$a$", therefore the whole space has some degree of indeterminacy. Neutrosophically, we write $a(0, 1, 0)$, which means the element $a$ is 100% unknown.

2) *Indeterminate Space*
*(due to Partially Known Element).*

Given the set $M = \{3, 4, 9(0.7, 0.1, 0.3)\}$, we have two elements $3$ and $4$ which surely belong to $M$, and one writes them neutrosophically as $3(1, 0, 0)$ and $4(1, 0, 0)$, while the third element $9$ belongs only partially ($70\%$) to $M$, its appurtenance to $M$ is indeterminate ($10\%$), and does not belong to $M$ (in a percentage of $30\%$).

Suppose the above neutrosophic set $M$ is endowed with a neutrosophic law * defined in the following way:

$$x_1(t_1, i_1, f_1) * x_2(t_2, i_2, f_2) = \max\{x_1, x_2\}( \min\{t_1, t_2\}, \max\{i_1, i_2\}, \max\{f_1, f_2\}), \quad (67)$$

which is a neutrosophic commutative semigroup with unit element $3(1, 0 ,0)$.

Clearly, if $x, y \in M$, then $x*y \in M$. Hence the neutrosophic law * is well defined.

Since *max* and *min* operators are commutative and associative, then * is also commutative and associative.

If $x \in M$, then $x*x = x$.

Below, examples of applying this neutrosophic law *:

3*9(0.7, 0.1, 0.3) = 3(1, 0, 0)*9(0.7, 0.1, 0.3) = max{3, 9}( min{1, 0.7}, max{0, 0.1}, max{0, 0.3} ) = 9(0.7, 0.1, 0.3).

3*4 = 3(1, 0, 0)*4(1, 0, 0) = max{3, 4}( min{1, 1}, max{0, 0}, max{0, 0} ) = 4(1, 0, 0).

2) *Indeterminate Law (Operation).*

For example, let the set (space) be NG = ( {0, 1, 2}, / ), where "/" means division.

NG is a *(t, i, f)-neutrosophic groupoid*, because the operation "/" (division) is partially defined, partially indeterminate (undefined), and partially not defined. Undefined is different from not defined. Let's see:

*2/1 = 1*, which belongs to *NG*; {defined}.

*1/0* = undefined; {indeterminate}.

*1/2 = 0.5*, which does not belongs to *NG*; {not defined}.

So the law defined on the set *NG* has the properties that:

- applying this law to some elements, the results are in *NG* [well defined law];
- applying this law to other elements, the results are not in *NG* [not well defined law];
- applying this law to again other elements, the results are undefined [indeterminate law].

We can construct many such algebraic structures where at least one axiom has such behavior (such indeterminacy in principal).

## 4.14 Websites at UNM for Neutrosophic Algebraic Structures and respectively Neutrosophic Geometries.

*http://fs.gallup.unm.edu/neutrosophy.htm*
and
*http://fs.gallup.unm.edu/geometries.htm*
respectively.

## 4.15 Acknowledgement.

The author would like to thank Mr. Mumtaz Ali, from Quaid-i-Azam University, Islamabad, Pakistan, Mr. Said Broumi, from University of Hassan II Mohammedia, Casablanca, Morocco, and Dr. W. B. Vasantha Kandasamy from Indian Institute of Technology, Chennai, Tamil Nadu, India, for their comments on this paper.

## 4.16 References.

## II. Neutrosophic Sets and Systems

## III. Neutrosophic Geometries

## IV. Refined Neutrosophics

76. Florentin Smarandache, *n-Valued Refined Neutrosophic Logic and Its Applications in Physics*, Progress in Physics, USA, 143-146, Vol. 4, 2013.

# 5 Refined Literal Indeterminacy and the Multiplication Law of Subindeterminacies

## 5.1 Abstract.

In this chapter, we make a short history of: the *neutrosophic set, neutrosophic numerical components and neutrosophic literal components, neutrosophic numbers, neutrosophic intervals, neutrosophic dual number, neutrosophic special dual number, neutrosophic special quasi dual number, neutrosophic quaternion number, neutrosophic octonion number, neutrosophic linguistic number, neutrosophic linguistic interval-style number, neutrosophic hypercomplex numbers of dimension* n, *and elemen-tary neutrosophic algebraic structures*. Afterwards, their generalizations to *refined neutrosophic set*, respectively *refined neutrosophic numerical and literal components*, then *refined neutrosophic numbers and refined neutrosophic algebraic structures*, and *set-style neutrosophic numbers*.

The aim of this chapter is to construct examples of splitting the literal indeterminacy *(I)* into literal sub-indeterminacies $(I_1, I_2, \ldots, I_r)$, and to define a *multiplication law* of these literal sub-indeterminacies in order to be able to build refined *I-neutrosophic algebraic structures*. Also, we give examples of splitting the numerical indeterminacy *(i)* into numerical sub-indeterminacies, and examples of splitting neutrosophic numerical components into neutrosophic numerical sub-components.

## 5.2 Introduction.

Neutrosophic Set was introduced in 1995 by Florentin Smarandache, who coined the words „neutrosophy” and its derivative „neutrosophic”. The first published work on neutrosophics was in 1998 {see [1]}.

There exist two types of neutrosophic components: numerical and literal.

## 5.3 Neutrosophic Numerical Components.

Of course, the *neutrosophic numerical components* $(t, i, f)$ are crisp numbers, intervals, or in general subsets of the unitary standard or nonstandard unit interval.

Let $\mathcal{U}$ be a universe of discourse, and $M$ a set included in $\mathcal{U}$. A generic element $x$ from $\mathcal{U}$

belongs to the set $M$ in the following way: $x(t, i, f) \in M$, meaning that $x$'s degree of membership/truth with respect to the set $M$ is $t$, $x$'s degree of indeterminacy with respect to the set $M$ is $i$, and $x$'s degree of non-membership/falsehood with respect to the set $M$ is $f$, where $t, i, f$ are independent standard subsets of the interval $[0,1]$, or non-standard subsets of the non-standard interval $]^{-}0, 1^{+}[$ in the case when one needs to make distinctions between *absolute and relative* truth, indeterminacy, or falsehood.

Many papers and books have been published for the cases when $t, i, f$ were single values (crisp numbers), or $t, i, f$ were intervals.

## 5.4 Neutrosophic Literal Components.

In 2003, W. B. Vasantha Kandasamy and Florentin Smarandache [4] introduced the *literal indeterminacy* "$I$", such that $I^2 = I$ (whence $I^n = I$ for $n \geq 1$, $n$ integer).

They extended this to *neutrosophic numbers* of the form: $a + bI$, where $a, b$ are real or complex numbers, and

$$(a_1 + b_1 I) + (a_2 + b_2 I) = (a_1 + a_2) + (b_1 + b_2)I \quad (68)$$

$$(a_1 + b_1 I)(a_2 + b_2 I) = (a_1 a_2) + (a_1 b_2 + a_2 b_1 + b_1 b_2)I \quad (69)$$

and developed many $I$-neutrosophic algebraic structures based on sets formed of neutrosophic numbers.

Working with imprecisions, Kandasamy & Smarandache have proposed (approximated) $I^2$ *by* $I$; yet different approaches may be investigated by the interested researchers where $I^2 \neq I$ (in accordance with their believe and with the practice), and thus a new field would arise in the neutrosophic theory.

The neutrosophic number $N = a + bI$ can be interpreted as: "$a$" represents the determinate part of number $N$, while "$bI$" the indeterminate part of number $N$, where indeterminacy $I$ may belong to a known (or unknown) set (not necessarily interval).

For example, $\sqrt{7} = 2.6457...$ that is irrational has infinitely many decimals. We cannot work with this exact number in our real life, we need to approximate it. Hence, we may write it as $2 + I$ with $I \in (0.6, 0.7)$, or as $2.6 + 3I$ with $I \in (0.01, 0.02)$, or $2.64 + 2I$ with $I \in (0.002, 0.004)$, etc. depending on the problem to be solved and on the needed accuracy.

Jun Ye [9] applied the neutrosophic numbers to decision making in *2014*.

The neutrosophic number $a+bI$ can be extended to a *Set-Style Neutrosophic Number A+BI*,

where $A$ and $B$ are sets, while $I$ is indeterminacy. As an interesting particular case one has when $A$ and $B$ are intervals, which is called *Interval-Style Neutrosophic Number.*

For example, {2, 3, 5} + {0, 4, 8, 12}I, with $I \in (0.5, 0.9)$, is a set-style neutrosophic number.

While $[30, 40] + [-10, -20]I$, with $I \in [7, 14]$, is an interval-style neutrosophic number.

## 5.5 Generalized Neutrosophic Complex Numbers

For a *generalized neutrosophic complex number*, which has the form $N = (a+bI_1) + (c+dI_2)i$, where $i = \sqrt{-1}$, one has $I_1$ = the indeterminacy of the real part of $N$, while $I_2$ = indeterminacy of the complex part of N. In particular cases we may have $I_1 = I_2$.

## 5.6 Neutrosophic Dual Numbers

A *dual number* [13] is a number

$$D = a + bg, \tag{70}$$

where $a$ and $b$ are real numbers, while $g$ is an element such that $g^2 = 0$.

Then, a *neutrosophic dual number*

$$ND = (a_0+a_1I_1) + (b_1+b_2I_2)g \tag{71}$$

where $a_0$, $a_1$, $b_1$, $b_2$ are real numbers, $I_1$ and $I_2$ are subindeterminacies, and $g$ is an element such that $g^2 = 0$.

A *dual number of dimension n* has the form

$$D_n = a_0 + b_1g_1+b_2g_2+\ldots+b_{n-1}g_{n-1} \tag{72}$$

where $a_0, b_1, b_2, \ldots, b_{n-1}$ are real numbers, while all $g_j$ are elements such that $g_j^2 = 0$ and $g_j\, g_k = g_k\, g_j = 0$ for all $j \neq k$.

One can generalize this to a *dual complex number of dimension n*, considering the same definition as (5), but taking $a_0, b_1, b_2, \ldots, b_{n-1}$ as complex numbers.

Now, a *neutrosophic dual number of dimension n* has the form:

$$ND_n = (a_{00}+a_{01}I_0) + (b_{11}+b_{12}I_1)g_1 + (b_{21}+b_{22}I_2)g_2 + \ldots + (b_{n-1,1}+b_{n-1,2}I_{n-1})g_{n-1} \tag{73}$$

where $a_{00}, a_{01}$, and all $b_{jk}$ are real or complex numbers, while $I_0, I_1, \ldots, I_{n-1}$ are subindeterminacies.

Similarly for *special dual numbers*, introduced by W. B. Vasantha & F. Smarandache [14], i.e. numbers of the form:

$$SD = a + bg, \tag{74}$$

where $a$ and $b$ are real numbers, while $g$ is an element such that $g^2 = g$ [for dimension $n$ one has $g_jg_k = g_kg_j = 0$ for $j \neq k$]; to observe that $g \neq I =$ indeterminacy, and in general the product of subindeterminacies

$$I_jI_k \neq 0 \text{ for } j \neq k], \tag{75}$$

and *special quasi dual number*, introduced by Vasantha-Smarandache [15], having the definition:

$$SQD = a + bg, \quad (76)$$

where $a$ and $b$ are real numbers, while $g$ is an element such that $g^2 = -g$ [for dimension $n$ one also has

$$g_j g_k = g_k g_j = 0 \text{ for } j \neq k], \quad (77)$$

and their corresponding forms for dimension $n$.

They all can be extended to *neutrosophic special dual number* and respectively *neutrosophic special quasi dual number* (of dimension $2$, and similarly for dimension $n$) in a same way.

### 5.6.1 Neutrosophic Quaternion Number.

A *quaternion number* is the number of the form:

$$H = a \cdot 1 + b \cdot i + c \cdot j + d \cdot k, \quad (78)$$

where

$$i^2 = j^2 = k^2 = i \cdot j \cdot k = -1, \quad (79)$$

and $a, b, c, d$ are real numbers.

A **neutrosophic quaternion number** is a number of the form:

$$NH = (a_1+a_2I) \cdot 1 + (b_1+b_2) \cdot i + (c_1+c_2I) \cdot j + (d_1+d_2I) \cdot k, \quad (80)$$

where $a_1, a_2, b_1, b_2, c_1, c_2, d_1, d_2$ are real or complex numbers, and $I$ = indeterminacy.

See: Weisstein, Eric W. "Quaternion." From *MathWorld* --A Wolfram Web Resource. http://mathworld.wolfram.com/Quaternion.html

### 5.6.2 Neutrosophic Octonion Number.

An *octonion number* has the form:

$$O = a + b_0i_0 + b_1i_1 + b_2i_2 + b_3i_3 + b_4i_4 + b_5i_5 + b_6i_6, \quad (81)$$

where $a, b_0, b_1, b_2, b_3, b_4, b_5, b_6$ are real numbers, and each of the triplets $(i_0, i_1, i_3)$, $(i_1, i_2, i_4)$, $(i_2, i_3, i_5)$, $(i_3, i_4, i_6)$, $(i_4, i_5, i_0)$, $(i_5, i_6, i_1)$, $(i_6, i_0, i_2)$ bears like the quaternions $(i, j, k)$.

A **neutrosophic octonion number** has the form:

$$NO = (a_1+a_2I) + (b_{01} + b_{02}I)i_0 + (b_{11} + b_{12}I)i_1 + (b_{21} + b_{22}I)i_2 + (b_{31} + b_{32}I)i_3 + (b_{41} + b_{42}I)i_4 + (b_{51} + b_{52}I)i_5 + (b_{61} + b_{62}I)i_6 \quad (82)$$

where all $a_1, a_2, b_{01}, b_{02}, b_{11}, b_{12}, b_{21}, b_{22}, b_{31}, b_{32}, b_{41}, b_{42}, b_{51}, b_{52}, b_{61}, b_{62}$ are real or complex numbers, $I$ = indeterminacy, and each of the triplets $(i_0, i_1, i_3)$, $(i_1, i_2, i_4)$, $(i_2, i_3, i_5)$, $(i_3, i_4, i_6)$, $(i_4, i_5, i_0)$, $(i_5, i_6, i_1)$, $(i_6, i_0, i_2)$ bears like the quaternions $(i, j, k)$.

See: Weisstein, Eric W. "Octonion." From *MathWorld* --A Wolfram Web Resource. http://mathworld.wolfram.com/Octonion.html

## 5.7 Neutrosophic Linguistic Numbers

A *neutrosophic linguistic number* has the shape*:*

$$N = L_{j+aI}, \quad (83)$$

where "L" means label or instance of a linguistic variable

$V = \{L_0, L_1, L_2, \ldots, L_p\}$, with $p \geq 1$, (84)

$j$ is a positive integer between $0$ and $p$-$1$, $a$ is a real number, and $I$ is indeterminacy that belongs to some real set, such that

$0 \leq min\{j+aI\} \leq max\{j+aI\} \leq p.$ (85)

*Neutrosophic linguistic interval-style number* has the form:

$N = [L_{j+aI}, L_{k+bI}]$ (86)

with similar restrictions (5) for $L_{k+bI}$.

## 5.8 Neutrosophic Intervals

We now for the first time extend the neutrosophic number to (open, closed, or half-open half-closed) neutrosophic interval.

A *neutrosophic interval A* is an (open, closed, or half-open half-closed) interval that has some indeterminacy in one of its extremes, i.e. it has the form $A = [a, b] \cup \{cI\}$, or $A = \{cI\} \cup [a, b]$, where $[a, b]$ is the determinate part of the neutrosophic interval A, and $I$ is the indeterminate part of it (while $a, b, c$ are real numbers, and $\cup$ means union). (Herein $I$ is an interval.)

We may even have neutrosophic intervals with double indeterminacy (or refined indeter-minacy): one to the left ($I_1$), and one to the right ($I_2$):

$$A = \{c_1 I_1\} \cup [a, b] \cup \{c_2 I_2\}. \tag{87}$$

**A classical real interval that has a neutrosophic number as one of its extremes becomes a neutrosophic interval. For example: *[0, $\sqrt{7}$]* can be represented as *[0, 2]*** $\cup$ *I* with $I = (2.0, 2.7)$, or ***[0, 2]*** $\cup$ *{10I}* with $I = (0.20, 0.27)$, or ***[0, 2.6]*** $\cup$ *{10I}* with $I = (0.26, 0.27)$, or ***[0, 2.64]*** $\cup$ *{10I}* with $I = (0.264, 0.265)$, etc. in the same way depending on the problem to be solved and on the needed accuracy.

We gave examples of closed neutrosophic intervals, but the open and half-open half-closed neutrosophic intervals are similar.

## 5.9 Notations

In order to make distinctions between the numerical and literal neutrosophic components, we start denoting the *numerical indeterminacy* by lower case letter "*i*" (whence consequently similar notations for *numerical truth* "*t*", and for *numerical falsehood* "*f*"), and *literal indeterminacy* by upper case letter "*I*" (whence consequently similar notations for *literal truth* "*T*", and for *literal falsehood* "*F*").

## 5.10 Refined Neutrosophic Components

In *2013*, F. Smarandache [3] introduced the refined neutrosophic components in the following way: the neutrosophic numerical components $t, i, f$ can be refined (split) into respectively the following refined neutrosophic numerical sub-components:

$$\langle t_1, t_2, \dots t_p;\ i_1, i_2, \dots i_r;\ f_1, f_2, \dots f_s; \rangle, \qquad (88)$$

where $p, r, s$ are integers $\geq 1$ and $\max\{p, r, s\} \geq 2$, meaning that at least one of $p, r, s$ is $\geq 2$; and $t_j$ represents types of numeral truths, $i_k$ represents types of numeral indeterminacies, and $f_l$ represents types of numeral falsehoods, for $j = 1, 2, \dots, p$; $k = 1, 2, \dots, r$; $l = 1, 2, \dots, s$.

$t_j, i_k, f_l$ are called numerical subcomponents, or respectively *numerical sub-truths*, *numerical sub-indeterminacies*, and *numerical sub-falsehoods*.

Similarly, the neutrosophic literal components $T, I, F$ can be refined (split) into respectively the following neutrosophic literal subcomponents:

$$\langle T_1, T_2, \dots T_p;\ I_1, I_2, \dots I_r;\ F_1, F_2, \dots F_s; \rangle, \qquad (89)$$

where $p, r, s$ are integers $\geq 1$ too, and $\max\{p, r, s\} \geq 2$, meaning that at least one of $p, r, s$ is $\geq 2$; and similarly $T_j$ represent types of literal truths, $I_k$ represent types of literal indeterminacies, and $F_l$ represent types of literal falsehoods, for $j = 1, 2, \dots, p$; $k = 1, 2, \dots, r$; $l = 1, 2, \dots, s$.

$T_j, I_k, F_l$ are called literal subcomponents, or respectively *literal sub-truths*, *literal sub-indeterminacies*, and *literal sub-falsehoods*.

Let consider a *simple example of refined numerical components*.

Suppose that a country $C$ is composed of two districts $D_1$ and $D_2$, and a candidate John Doe competes for the position of president of this country $C$. Per whole country, $NL$ (Joe Doe) = $(0.6, 0.1, 0.3)$, meaning that 60% of people voted for him, 10% of people were indeterminate or neutral – i.e. didn't vote, or gave a black vote, or a blank vote –, and 30% of people voted against him, where $NL$ means the neutrosophic logic values.

But a political analyst does some research to find out what happened to each district separately. So, he does a refinement and he gets:

$$\begin{pmatrix} 0.40 & 0.20 \\ t_1 & t_2 \end{pmatrix}; \begin{pmatrix} 0.08 & 0.02 \\ i_1 & i_2 \end{pmatrix}; \begin{pmatrix} 0.05 & 0.25 \\ f_1 & f_2 \end{pmatrix} \quad (90)$$

which means that 40% of people that voted for Joe Doe were from district $D_1$, and 20% of people that voted for Joe Doe were from district $D_2$; similarly, 8% from $D_1$ and 2% from $D_2$ were indeterminate (neutral), and 5% from $D_1$ and 25% from $D_2$ were against Joe Doe.

It is possible, in the same example, to refine (split) it in a different way, considering another criterion, namely: what percentage of people did not vote ($i_1$), what percentage of people gave a blank vote – cutting all candidates on the ballot – ($i_2$), and what percentage of people gave a blank vote – not selecting any candidate on the ballot ($i_3$). Thus, the numerical indeterminacy ($i$) is refined into $i_1$, $i_2$, and $i_3$:

$$\begin{pmatrix} \frac{0.60}{t}; & \frac{0.05}{i_1} & \frac{0.04}{i_2} & \frac{0.01}{i_3}; & \frac{0.30}{f} \end{pmatrix} \quad (91)$$

## 5.11 Refined Neutrosophic Numbers

In 2015, F. Smarandache [6] introduced the *refined literal indeterminacy* ($I$), which was split (refined) as $I_1, I_2, \dots, I_r$, with $r \geq 2$, where $I_k$, for $k = 1, 2, \dots, r$ represent types of literal sub-indeterminacies. A refined neutrosophic number has the general form:

$$N_r = a + b_1 I_1 + b_2 I_2 + \cdots + b_r I_r, \quad (92)$$

where $a, b_1, b_2, \dots, b_r$ are real numbers, and in this case $N_r$ is called a *refined neutrosophic real number*; and if at least one of $a, b_1, b_2, \dots, b_r$ is a complex number (i.e. of the form $\alpha + \beta\sqrt{-1}$, with $\beta \neq 0$, and α, β real numbers ), then $N_r$ is called a *refined neutrosophic complex number*.

An example of refined neutrosophic number, with three types of indeterminacies resulted from the cubic root ($I_1$), from Euler's constant $e$ ($I_2$), and from number $\pi$ ($I_3$):

$$N_3 = -6 + \sqrt[3]{59} - 2e + 11\pi \qquad (93)$$

Roughly,

$N_3 = -6 + (3 + I_1) - 2(2 + I_2) + 11(3 + I_3)$

$= (-6 + 3 - 4 + 33) + I_1 - 2I_2 + 11I_3 = 26 + I_1 - 2I_2 + 11I_3$

where $I_1 \in (0.8, 0.9)$, $I_2 \in (0.7, 0.8)$, and $I_3 \in (0.1, 0.2)$, since $\sqrt[3]{59} = 3.8929\ldots$, $e = 2.7182\ldots$, $\pi = 3.1415\ldots$ .

Of course, other *3*-valued refined neutrosophic number representations of $N_3$ could be done depending on accuracy.

Then F. Smarandache [6] defined the *refined I-neutrosophic algebraic structures* in 2015 as algebraic structures based on sets of refined neutrosophic numbers.

Soon after this definition, Dr. Adesina Agboola wrote a paper on refined *I*-neutrosophic algebraic structures [7].

They were called "*I*-neutrosophic" because the refinement is done with respect to the literal indeterminacy ($I$), in order to distinguish them from the refined $(t, i, f)$-neutrosophic algebraic structures, where " $(t, i, f)$ -neutrosophic" is referred to as refinement of the neutrosophic numerical components $t, i, f$.

Said Broumi and F. Smarandache published a paper [8] on refined neutrosophic numerical components in 2014.

## 5.12 Neutrosophic Hypercomplex Numbers of Dimension *n*

The *Hypercomplex Number of Dimension n* (or *n-Complex Number*) was defined by S. Olariu [10] as a number of the form:

$$u = x_o + h_1 x_1 + h_2 x_2 + \ldots + h_{n-1} x_{n-1} \quad (94)$$

where $n \geq 2$, and the variables $x_0, x_1, x_2, \ldots, x_{n-1}$ are real numbers, while $h_1, h_2, \ldots, h_{n-1}$ are the complex units, $h_o = 1$, and they are multiplied as follows:

$$h_j h_k = h_{j+k} \text{ if } 0 \leq j+k \leq n-1, \text{ and } h_j h_k = h_{j+k-n} \text{ if } n \leq j+k \leq 2n-2. \quad (95)$$

We think that the above (11) complex unit multiplication formulas can be written in a simpler way as:

$$h_j h_k = h_{j+k \, (mod\, n)} \quad (96)$$

where *mod n* means *modulo n.* For example, if $n = 5$, then $h_3 h_4 = h_{3+4(mod\, 5)} = h_{7(mod 5)} = h_2$.

Even more, formula above allows us to multiply many complex units at once, as follows:

$h_{j1}h_{j2}\ldots h_{jp} = h_{j1+j2+\ldots+jp\,(mod\,n)}$, *for* $p \geq 1$. (97)

We now define for the first time the *Neutrosophic Hypercomplex Number of Dimension n* (or *Neutrosophic n-Complex Number*), which is a number of the form:

$u+vI$, (98)

where $u$ and $v$ are $n$-complex numbers and $I$ = indeterminacy.

We also introduce now the *Refined Neutrosophic Hypercomplex Number of Dimension n* (or *Refined Neutrosophic n-Complex Number*) as a number of the form:

$u+v_1I_1+v_2I_2+\ldots+v_rI_r$ (99)

where $u, v_1, v_2, \ldots, v_r$ are $n$-complex numbers, and $I_1, I_2, \ldots, I_r$ are sub-indeterminacies, for $r \geq 2$.

Combining these, we may define a *Hybrid Neutrosophic Hypercomplex Number* (or *Hybrid Neutrosophic n-Complex Number*), which is a number of the form $u+vI$, where either $u$ or $v$ is a $n$-complex number while the other one is different (may be an $m$-complex number, with $m \neq n$, or a real number, or another type of number).

And a *Hybrid Refined Neutrosophic Hypercomplex Number* (or *Hybrid Refined Neutrosophic n-Complex Number*), which is a number of the form $u+v_1I_1+v_2I_2+\ldots+v_rI_r$, where at least one of $u, v_1, v_2, \ldots, v_r$ is a $n$-complex number, while the others are different (may be $m$-complex numbers, with $m \neq n$, and/or a real numbers, and/or other types of numbers).

## 5.13 Neutrosophic Graphs

We now introduce for the first time the general definition of a *neutrosophic graph* [12], which is a (directed or undirected) graph that has some indeterminacy with respect to its edges, or with respect to its vertexes (nodes), or with respect to both (edges and vertexes simultaneously). We have four main categories of neutrosophic graphs:

1) The $(t, i, f)$*-Edge Neutrosophic Graph.*

In such a graph, the connection between two vertexes $A$ and $B$, represented by edge $AB$:

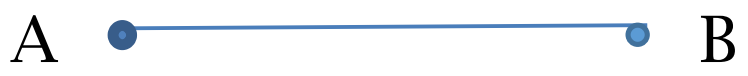

has the neutroosphic value of $(t, i, f)$.

2) $I$*-Edge Neutrosophic Graph.*

This one was introduced in 2003 in the book “Fuzzy Cognitive Maps and Neutrosophic Cognitive Maps”, by Dr. Vasantha Kandasamy and F. Smarandache, that used a different approach for the edge:

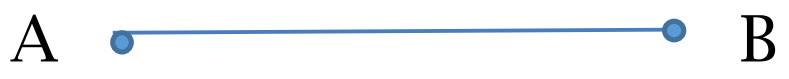

which can be just $I$ = literal indeterminacy of the edge, with $I^2 = I$ (as in $I$-Neutrosophic algebraic structures). Therefore, simply we say that the connection between vertex $A$ and vertex $B$ is indeterminate.

3) *Orientation-Edge Neutrosophic Graph*.

At least one edge, let's say AB, has an unknown orientation (i.e. we do not know if it is from A to B, or from B to A).

4) $I$*-Vertex Neutrosophic Graph.*

Or at least one literal indeterminate vertex, meaning we do not know what this vertex represents.

5) $(t, i, f)$*-Vertex Neutrosophic Graph.*

We can also have at least one neutrosophic vertex, for example vertex $A$ only partially belongs to the graph $(t)$, indeterminate appurtenance to the graph $(i)$, does not partially belong to the graph $(f)$, we can say $A(t, i, f)$.

And combinations of any two, three, four, or five of the above five possibilities of neutrosophic graphs.

If $(t, i, f)$ or the literal $I$ are refined, we can get corresponding *refined neurosophic graphs*.

## 5.14 Example of Refined Indeterminacy and Multiplication Law of Subindeterminacies

Discussing the development of Refined $I$-Neutrosophic Structures with Dr. W.B. Vasantha Kandasamy, Dr. A.A.A. Agboola, Mumtaz Ali, and Said Broumi, a question has arisen: if $I$ is refined into $I_1, I_2, \dots, I_r$, with $r \geq 2$, how to define (or compute) $I_j * I_k$, for $j \neq k$?

We need to design a Sub-Indeterminacy $*$ Law Table.

Of course, this depends on the way one defines the algebraic binary multiplication law $*$ on the set:

$$\{N_r = a + b_1 I_1 + b_2 I_2 + \dots + b_r I_r | a, b_1, b_2, \dots, b_r \in M\}, \quad (100)$$

where $M$ can be $\mathbb{R}$ (the set of real numbers), or $\mathbb{C}$ (the set of complex numbers).

We present the below example.

But, first, let's present several (possible) interconnections between logic, set, and algebra.

| | *Logic* | *Set* | *Algebra* |
|---|---|---|---|
| *operators* | Disjunction (or) ∨ | Union ∪ | Addition + |
| | Conjunction (and) ∧ | Intersection ∩ | Multiplication · |
| | Negation ¬ | Complement ∁ | Subtraction − |
| | Implication → | Inclusion ⊆ | Subtraction, Addition −, + |
| | Equivalence ↔ | Identity ≡ | Equality = |

*Table 1: Interconnections between logic, set, and algebra.*

In general, if a Venn Diagram has $n$ sets, with $n \geq 1$, the number of disjoint parts formed is $2^n$. Then, if one combines the $2^n$ parts either by none, or by one, or by 2, …, or by $2^n$, one gets:

$$C_{2^n}^0 + C'_{2^n} + C_{2^n}^2 + \cdots + C_{2^n}^{2^n} = (1+1)^{2^n} = 2^{2^n}. \quad (101)$$

Hence, for $n = 2$, the Venn Diagram, with literal truth ($T$), and literal falsehood ($F$), will make $2^2 = 4$ disjoint parts, where the whole rectangle represents the whole universe of discourse ($\mathcal{U}$).

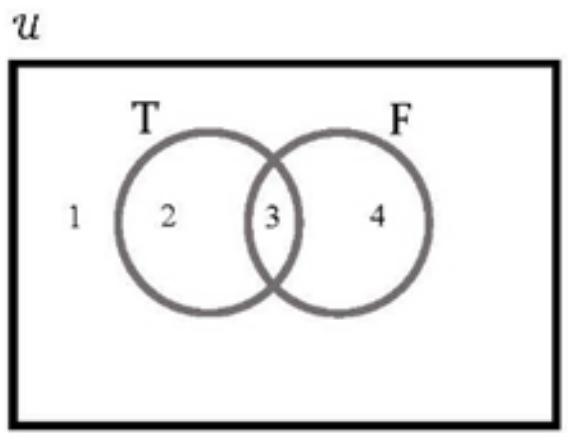

*Fig. 5 Venn Diagram for n =2.*

Then, combining the four disjoint parts by none, by one, by two, by three, and by four, one gets

$$C_4^0 + C_4^1 + C_4^2 + C_4^3 + C_4^4 = (1+1)^4 = 2^4 = 16 = 2^{2^2}. \quad (102)$$

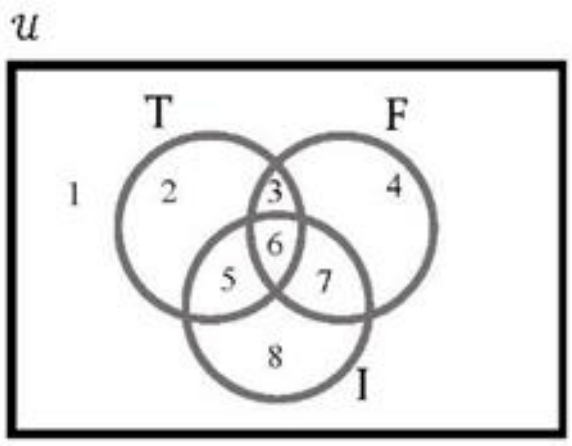

*Fig. 6 Venn Diagram for n = 3.*

For $n = 3$, one has $2^3 = 8$ disjoint parts, and combining them by none, by one, by two, and so on, by eight, one gets $2^8 = 256$, or $2^{2^3} = 256$.

For the case when $n = 2 = \{T, F\}$ one can make up to *16* sub-indeterminacies, such as:

$I_1 = C =$ **contradiction** $=$ True and False $= T \wedge F$

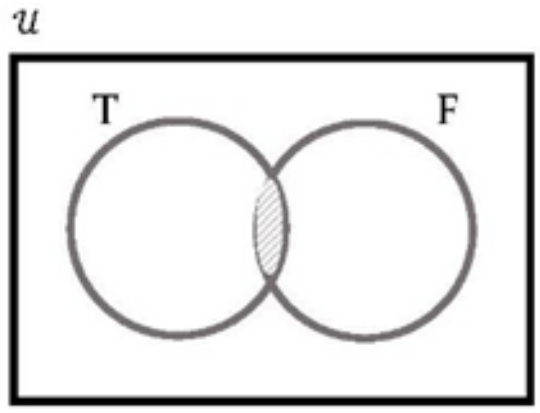

*Fig. 7*

$I_2 = Y =$ **uncertainty** $=$ True or False $= T \vee F$

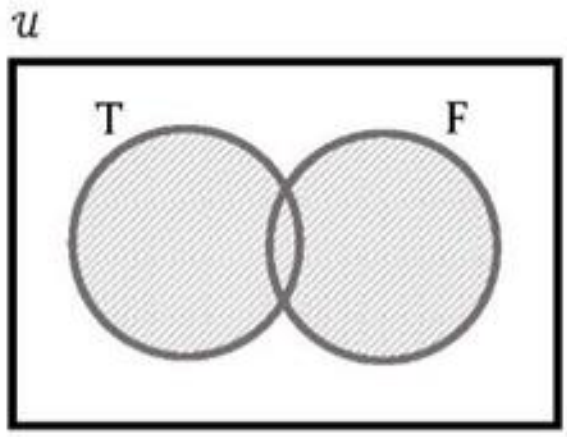

*Fig. 8*

$I_3 = S =$ **unsureness** $=$ either True or False $= T \underline{\vee} F$

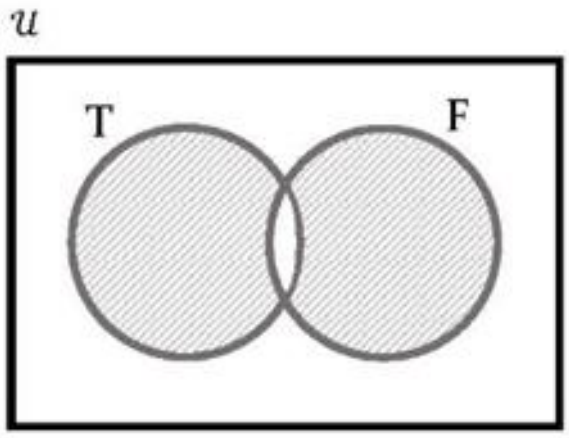

*Fig. 9*

$I_4 = H =$ **nihilness** = neither True nor False $= \neg T \wedge \neg F$

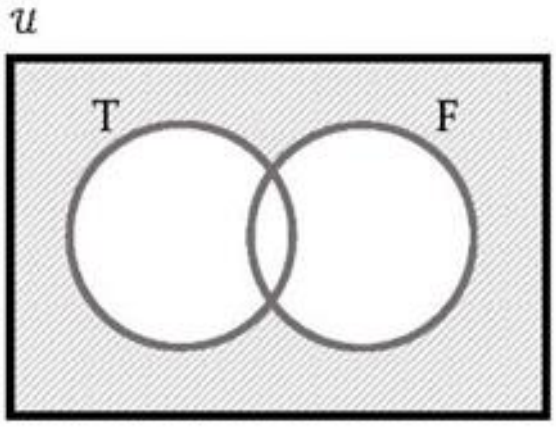

*Fig. 10*

$I_5 = V =$ **vagueness** = not True or not False $= \neg T \vee \neg F$

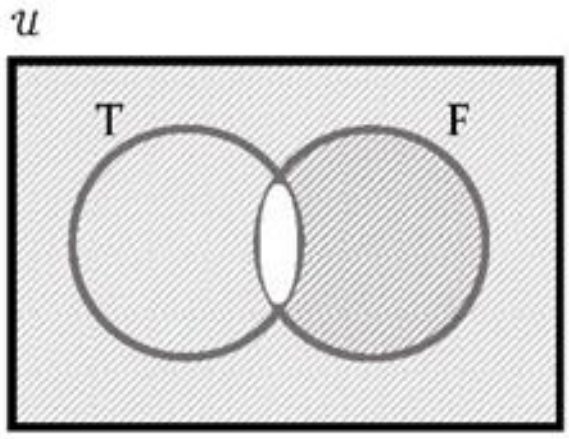

*Fig. 11*

$I_6 = E =$ **emptiness** = neither True nor not True

$= \neg T \wedge \neg(\neg T) = \neg T \wedge T$

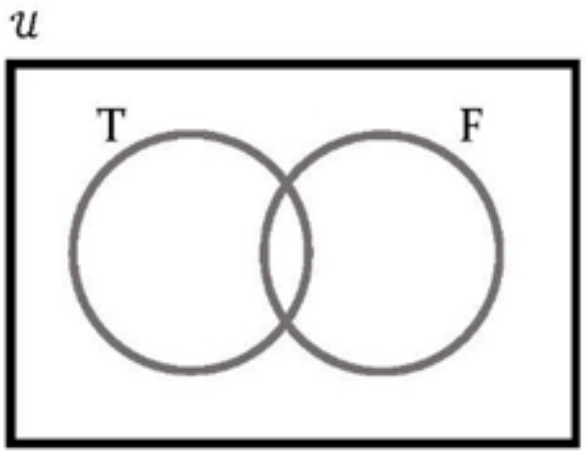

*Fig. 12*

Let's consider the literal indeterminacy ($I$) refined into only six literal sub-indeterminacies as above.

The binary multiplication law

$$*: \{I_1, I_2, I_3, I_4, I_5, I_6\}^2 \to \{I_1, I_2, I_3, I_4, I_5, I_6\} \quad (103)$$

defined as:

$I_j * I_k$ = intersections of their Venn diagram representations; or $I_j * I_k$ = application of $\wedge$ operator, i.e. $I_j \wedge I_k$.

We make the following:

| * | $I_1$ | $I_2$ | $I_3$ | $I_4$ | $I_5$ | $I_6$ |
|---|---|---|---|---|---|---|
| $I_1$ | $I_1$ | $I_1$ | $I_6$ | $I_6$ | $I_6$ | $I_6$ |
| $I_2$ | $I_1$ | $I_2$ | $I_3$ | $I_6$ | $I_3$ | $I_6$ |
| $I_3$ | $I_6$ | $I_3$ | $I_3$ | $I_6$ | $I_3$ | $I_6$ |
| $I_4$ | $I_6$ | $I_6$ | $I_6$ | $I_4$ | $I_4$ | $I_6$ |
| $I_5$ | $I_6$ | $I_3$ | $I_3$ | $I_4$ | $I_5$ | $I_6$ |
| $I_6$ | $I_6$ | $I_6$ | $I_6$ | $I_6$ | $I_6$ | $I_6$ |

*Table 2: Sub-Indeterminacies Multiplication Law*

## 5.15 Remark on the Variety of Sub-Indeterminacies Diagrams

One can construct in various ways the diagrams that represent the sub-indeterminacies and similarly one can define in many ways the $*$ algebraic multiplication law, $I_j * I_k$, depending on the problem or application to solve.

What we constructed above is just an example, not a general procedure.

Let's present below several calculations, so the reader gets familiar:

$I_1 * I_2 =$ (shaded area of $I_1$) $\cap$ (shaded area of $I_2$) $=$ shaded area of $I_1$,

or $I_1 * I_2 = (T \wedge F) \wedge (T \vee F) = T \wedge F = I_1$.

$I_3 * I_4 =$ (shaded area of $I_3$) $\cap$ (shaded area of $I_4$) $=$ empty set $= I_6$,

or $I_3 * I_4 = \left(T \underline{\vee} F\right) \wedge (\neg T \wedge \neg F) = [T \wedge (\neg T \wedge \neg F)] \underline{\vee} [F \wedge (\neg T \wedge \neg F)] = (T \wedge \neg T \wedge \neg F) \underline{\vee} (F \wedge \neg T \wedge \neg F) =$ (impossible) $\underline{\vee}$ (impossible)

because of $T \wedge \neg T$ in the first pair of parentheses and because of $F \wedge \neg F$ in the second pair of parentheses

$=$ (impossible) $= I_6$.

$I_5 * I_5 =$ (shaded area of $I_5$) $\cap$ (shaded area of $I_5$) $=$ (shaded area of $I_5$) $= I_5$,

or $I_5 * I_5 = (\neg T \vee \neg F) \wedge (\neg T \vee \neg F) = \neg T \vee \neg F = I_5$.

Now we are able to build refined $I$-neutrosophic algebraic structures on the set

$$S_6 = \{a_0 + a_1 I_1 + a_2 I_2 + \cdots + a_6 I_6, \text{for } a_0, a_1, a_2, \ldots\ a_6 \in \mathbb{R}\}, \qquad (104)$$

by defining the addition of refined I-neutrosophic numbers:

$$(a_0 + a_1 I_1 + a_2 I_2 + \cdots + a_6 I_6) + (b_0 + b_1 I_1 + b_2 I_2 + \cdots + b_6 I_6) = (a_0 + b_0) + (a_1 + b_1) I_1 + (a_2 + b_2) I_2 + \cdots + (a_6 + b_6) I_6 \in S_6. \quad (105)$$

And the multiplication of refined neutrosophic numbers:

$$(a_0 + a_1 I_1 + a_2 I_2 + \cdots + a_6 I_6) \cdot (b_0 + b_1 I_1 + b_2 I_2 + \cdots + b_6 I_6) = a_0 b_0 + (a_0 b_1 + a_1 b_0) I_1 + (a_0 b_2 + a_2 b_0) I_2 + \cdots + (a_0 b_6 + a_6 b_0) I_6 + + \sum_{j,k=1}^{6} a_j b_k \left(I_j * I_k\right) = a_0 b_0 + \sum_{k=1}^{6} (a_0 b_k + a_k b_0) I_k + \sum_{j,k=1}^{6} a_j b_k \left(I_j * I_k\right) \in S_6, \quad (106)$$

where the coefficients (scalars) $a_m \cdot b_n$ , for $m = 0, 1, 2, \dots, 6$ and $n = 0, 1, 2, \dots, 6$, are multiplied as any real numbers, while $I_j * I_k$ are calculated according to the previous Sub-Indeterminacies Multiplication Law (Table 2).

Clearly, both operators (addition and multiplication of refined neutrosophic numbers) are well-defined on the set $S_6$.

## 5.16 References.

# 6 Neutrosophic Actions, Prevalence Order, Refinement of Neutrosophic Entities, and Neutrosophic Literal Logical Operators

## 6.1 Abstract.

In this chapter, we define for the first time three *neutrosophic actions* and their properties. We then introduce the *prevalence order* on $\{T, I, F\}$ with respect to a given neutrosophic operator "$o$", which may be subjective - as defined by the neutrosophic experts. And the *refinement of neutrosophic entities* <A>, <neutA>, and <antiA>.

Then we extend the classical logical operators to *neutrosophic literal logical operators* and to *refined literal logical operators*, and we define the *refinement neutrosophic literal space*.

## 6.2 Introduction.

In Boolean Logic, a proposition $\mathcal{P}$ is either true (T), or false (F).

In Neutrosophic Logic, a proposition $\mathcal{P}$ is either true (T), false (F), or indeterminate (I).

For example, in Boolean Logic the proposition $\mathcal{P}_1$:

"1 + 1 = 2 (in base 10)"is true,

while the proposition $\mathcal{P}_2$:

"1 + 1 = 3 (in base 10)" is false.

In neutrosophic logic, besides propositions $\mathcal{P}_1$ (which is true) and $\mathcal{P}_2$ (which is false), we may also have proposition $\mathcal{P}_3$:

"1 + 1 = ? (in base 10)",

which is an incomplete/indeterminate proposition (neither true, nor false).

### 6.2.1 Remark.

All **conjectures in science** are indeterminate at the beginning (researchers not knowing if they are true or false), and later they are proved as being either true, or false, or indeterminate in the case they were unclearly formulated.

## 6.3 Notations.

In order to avoid confusions regarding the operators, we note them as:

Boolean (classical) logic:

$$\neg, \quad \wedge, \quad \vee, \quad \underline{\vee}, \quad \rightarrow, \quad \leftrightarrow$$

Fuzzy logic:

$$\underset{F}{\neg}, \quad \underset{F}{\wedge}, \quad \underset{F}{\vee}, \quad \underset{F}{\underline{\vee}}, \quad \underset{F}{\rightarrow}, \quad \underset{F}{\leftrightarrow}$$

Neutrosophic logic:

$$\neg_N, \quad \wedge_N, \quad \vee_N, \quad \underline{\vee}_N, \quad \rightarrow_N, \quad \leftrightarrow_N$$

## 6.4 Three Neutrosophic Actions.

In the frame of neutrosophy, we have considered [1995] for each entity $\langle A \rangle$, its opposite $\langle \text{anti}A \rangle$, and their neutrality $\langle \text{neut}A \rangle$ {i.e. neither $\langle A \rangle$, nor $\langle \text{anti}A \rangle$}. Also, by $\langle \text{non}A \rangle$ we mean what is not $\langle A \rangle$, i.e. its opposite $\langle \text{anti}A \rangle$, together with its neutral(ity) $\langle \text{neut}A \rangle$; therefore:

$$\langle \text{non}A \rangle = \langle \text{neut}A \rangle \vee \langle \text{anti}A \rangle.$$

Based on these, we may straightforwardly introduce for the first time the following neutrosophic actions with respect to an entity <A>:

1. **To neutralize** (or **to neuter**, or simply **to neut-ize**) the entity <A>. [As a noun: neutralization, or neuter-ization, or simply neut-ization.] We denote it by <neutA> or neut(A).
2. **To antithetic-ize** (or **to anti-ize**) the entity <A>. [As a noun: antithetic-ization, or anti-ization.] We denote it by <antiA> ot anti(A). This action is 100% opposition to entity <A> (strong opposition, or strong negation).

3. To **non-ize** the entity <A>. [As a noun: non-ization]. We denote it by <nonA> or non(A). It is an opposition in a percentage between (0, 100]% to entity <A> (weak opposition).

Of course, not all entities <A> can be neutralized, or antithetic-ized, or non-ized.

### 6.4.1 Example.

Let $\langle A \rangle$ = "Phoenix Cardinals beats Texas Cowboys".

Then,

$\langle \text{neut}A \rangle$
= "Phoenix Cardinals has a tie game with Texas Cowboys";
$\langle \text{anti}A \rangle$
= "Phoenix Cardinals is beaten by Texas Cowboys";
$\langle \text{non}A \rangle$
= "Phoenix Cardinals has a tie game with Texas Cowboys, or Phoenix Cardinals is beaten by Texas Cowboys".

### 6.4.2 Properties of the Three Neutrosophic Actions.

$$\text{neut}(\langle \text{anti}A \rangle) = \text{neut}(\langle \text{neut}A \rangle) = \text{neut}(A);$$

$$\text{anti}(\langle \text{anti}A \rangle) = A;\ \text{anti}(\langle \text{neut}A \rangle) = \langle A \rangle \text{ or } \langle \text{anti}A \rangle;$$

$$\text{non}(\langle \text{anti}A \rangle) = \langle A \rangle \text{ or } \langle \text{neut}A \rangle;\ \text{non}(\langle \text{neut}A \rangle) = \langle A \rangle \text{ or } \langle \text{anti}A \rangle.$$

## 6.5 Neutrosophic Actions' Truth-Value Tables.

Let's have a logical proposition P, which may be true (T), Indeterminate (I), or false (F) as in previous example. One applies the neutrosophic actions below.

### 6.5.1 Neutralization (or Indetermination) of P:

| **neut(P)** | **T** | **I** | **F** |
|---|---|---|---|
| | *I* | *I* | *I* |

*Table 3*

### 6.5.2 Antitheticization (Neutrosophic Strong Opposition to P):

| **anti(P)** | **T** | **I** | **F** |
|---|---|---|---|
| | $F$ | $T \vee F$ | $T$ |

*Table 4*

### 6.5.3 Non-ization (Neutrosophic Weak Opposition to P):

| **non(P)** | **T** | **I** | **F** |
|---|---|---|---|
| | $I \vee F$ | $T \vee F$ | $T \vee I$ |

*Table 5*

## 6.6 Refinement of Entities in Neutrosophy.

In neutrosophy, an entity $\langle A \rangle$ has an opposite $\langle \text{anti}A \rangle$ and a neutral $\langle \text{neut}A \rangle$.

But these three categories can be refined in sub-entities $\langle A \rangle_1, \langle A \rangle_2, \dots, \langle A \rangle_m$, and respectively $\langle \text{neut}A \rangle_1, \langle \text{neut}A \rangle_2, \dots, \langle \text{neut}A \rangle_n$, and also $\langle \text{anti}A \rangle_1, \langle \text{anti}A \rangle_2, \dots, \langle \text{anti}A \rangle_p$, where *m, n, p* are integers $\geq 1$, but $m + n + p \geq 4$ (meaning that at least one of $\langle A \rangle$, $\langle \text{anti}A \rangle$ or $\langle \text{neut}A \rangle$ is refined in two or more sub-entities).

For example, if $\langle A \rangle = \text{white color}$, then

$$\langle \text{anti}A \rangle = \text{black color},$$

while $\langle \text{neut}A \rangle = \text{colors different from white and black}$.

If we refine them, we get various nuances of white color: $\langle A \rangle_1, \langle A \rangle_2, \dots$, and various nuances of black color: $\langle \text{anti}A \rangle_1, \langle \text{anti}A \rangle_2, \dots$, and the colors in between them (red, green, yellow, blue, etc.): $\langle \text{neut}A \rangle_1, \langle \text{neut}A \rangle_2, \dots$ .

Similarly as above, we want to point out that not all entities <A> and/or their corresponding (if any) <neutA> and <antiA> can be refined.

## 6.7 The Prevalence Order.

Let's consider the classical literal (symbolic) truth *(T)* and falsehood *(F)*.

In a similar way, for neutrosophic operators we may consider the literal (symbolic) truth *(T)*, the literal (symbolic) indeterminacy *(I)*, and the literal (symbolic) falsehood *(F)*.

We also introduce the *prevalence order* on $\{T, I, F\}$ with respect to a given binary and commutative neutrosophic operator "$o$".

The neutrosophic operators are: neutrosophic negation, neutrosophic conjunction, neutrosophic disjunction, neutrosophic exclusive disjunction, neutrosophic Sheffer's stroke, neutrosophic implication, neutrosophic equivalence, etc.

The prevalence order is *partially objective* (following the classical logic for the relationship between *T* and *F*), and *partially subjective* (when the indeterminacy *I* interferes with itself or with *T* or *F*).

For its subjective part, the prevalence order is determined by the neutrosophic logic expert in terms of the application/problem to solve, and also depending on the specific conditions of the application/problem.

For $X \neq Y$, we write $X{\rm ℗}Y$, or $X \succ_o Y$, and we read "X" prevails to Y with respect to the neutrosophic binary commutative operator "o", which means that $XoY = X$.

Let's see the below examples. We mean by "o": conjunction, disjunction, exclusive disjunction, Sheffer's stroke, and equivalence.

## 6.8 Neutrosophic Literal Operators & Neutrosophic Numerical Operators.

1. If we mean by *neutrosophic literal proposition*, a proposition whose truth-value is a letter: either T or I or F. The operators that deal with such logical propositions are called *neutrosophic literal operators*.
2. And by *neutrosophic numerical proposition*, a proposition whose truth value is a triple of numbers (or in general of numerical subsets of the interval [0, 1]), for examples A(0.6, 0.1, 0.4) or B([0, 0.2], {0.3, 0.4, 0.6}, (0.7, 0.8)). The operators that deal with such logical propositions are called *neutrosophic numerical operators*.

## 6.9 Truth-Value Tables of Neutrosophic Literal Operators.

In Boolean Logic, one has the following truth-value table for negation:

### 6.9.1 Classical Negation.

| ¬ | T | F |
|---|---|---|
| | *F* | *T* |

*Table 6*

In Neutrosophic Logic, one has the following neutrosophic truth-value table for the neutrosophic negation:

### 6.9.2 Neutrosophic Negation.

| $\neg_N$ | T | I | F |
|---|---|---|---|
| | ◯*F* | *I* | ◯*T* |

*Table 7*

So, we have to consider that the negation of *I* is *I*, while the negations of *T* and *F* are similar as in classical logic.

In classical logic, one has:

### 6.9.3 Classical Conjunction.

| $\wedge$ | T | F |
|---|---|---|
| T | *T* | *F* |
| F | *F* | *F* |

*Table 8*

In neutrosophic logic, one has:

### 6.9.4 Neutrosophic Conjunction ($\boldsymbol{AND_N}$), version 1

| $\wedge_N$ | T | I | F |
|---|---|---|---|
| T | *T* | *I* | *F* |
| I | *I* | *I* | *I* |
| F | *F* | *I* | *F* |

*Table 9*

The objective part (circled literal components in the above table) remains as in classical logic, but when indeterminacy $I$ interferes, the neutrosophic expert may choose the most fit prevalence order.

There are also cases when the expert may choose, for various reasons, to entangle the classical logic in the objective part. In this case, the prevalence order will be totally subjective.

The prevalence order works for classical logic too. As an example, for classical conjunction, one has $F \succ_c T$, which means that $F \wedge T = F$.

While the prevalence order for the neutrosophic conjunction in the above tables was:

$$I \succ_c F \succ_c T, \quad (107)$$

which means that $I \wedge_N F = I$, and $I \wedge_N T = I$.

Other prevalence orders can be used herein, such as:

$$F \succ_c I \succ_c T, \quad (108)$$

and its corresponding table would be:

### 6.9.5 Neutrosophic Conjunction ($\boldsymbol{AND_N}$), version 2

| $\wedge_N$ | T | I | F |
|---|---|---|---|
| T | $T$ | $I$ | $F$ |
| I | $I$ | $I$ | $F$ |
| F | $F$ | $F$ | $F$ |

*Table 10*

which means that $F_{\wedge_N} I = F$ and $I_{\wedge_N} I = I$; or another prevalence order:

$$F \succ_c T \succ_c I, \qquad (109)$$

and its corresponging table would be:

### 6.9.6 Neutrosophic Conjunction ($\boldsymbol{AND_N}$), version 3

| $\wedge_N$ | T | I | F |
|---|---|---|---|
| T | $T$ | $T$ | $F$ |
| I | $T$ | $I$ | $F$ |
| F | $F$ | $F$ | $F$ |

*Table 11*

which means that $F_{\wedge_N} I = F$ and $T_{\wedge_N} I = T$.

If one compares the three versions of the neutrosophic literal conjunction, one observes that the objective part remains the same, but the subjective part changes.

The subjective of the prevalence order can be established in an optimistic way, or pessimistic way, or according to the weights assigned to the neutrosophic literal components T, I, F by the experts.

In a similar way, we do for disjunction.

In classical logic, one has:

### 6.9.7 Classical Disjunction.

| ∨ | T | F |
|---|---|---|
| T | $T$ | $T$ |
| F | $T$ | $F$ |

*Table 12*

In neutrosophic logic, one has:

### 6.9.8 Neutrosophic Disjunction ($\boldsymbol{OR_N}$)

| $\vee_N$ | T | I | F |
|---|---|---|---|
| T | $T$ | $T$ | $T$ |
| I | $T$ | $I$ | $F$ |
| F | $T$ | $F$ | $F$ |

*Table 13*

where we used the following prevalence order:

$$T \succ_d F \succ_d I, \tag{110}$$

but the reader is invited (as an exercise) to use another prevalence order, such as:

$$T \succ_d I \succ_d F, \tag{111}$$

or

$$I \succ_d T \succ_d F, \text{ etc.,} \tag{112}$$

for all neutrosophic logical operators presented above and below in this paper.

In classical logic, one has:

### 6.9.9 Classical Exclusive Disjunction

| $\underline{\vee}$ | T | F |
|---|---|---|
| T | *F* | *T* |
| F | *T* | *F* |

*Table 14*

In neutrosophic logic, one has:

### 6.9.10 Neutrosophic Exclusive Disjunction

| $\underline{\vee}_N$ | T | I | F |
|---|---|---|---|
| T | *F* | *T* | *T* |
| I | *T* | *I* | *F* |
| F | *T* | *F* | *F* |

*Table 15*

using the prevalence order

$$T \succ_d F \succ_d I. \quad (113)$$

In classical logic, one has:

### 6.9.11 Classical Sheffer's Stroke

| \| | T | F |
|---|---|---|
| T | *F* | *T* |
| F | *T* | *T* |

*Table 16*

In neutrosophic logic, one has:

### 6.9.12 Neutrosophic Sheffer's Stroke

| $\vert_N$ | T | I | F |
|---|---|---|---|
| T | *F* | *T* | *T* |
| I | *T* | *I* | *I* |
| F | *T* | *I* | *T* |

*Table 17*

using the prevalence order

$$T \succ_d I \succ_d F. \tag{114}$$

In classical logic, one has:

### 6.9.13 Classical Implication

| → | T | F |
|---|---|---|
| T | *T* | *F* |
| F | *T* | *T* |

*Table 18*

In neutrosophic logic, one has:

### 6.9.14 Neutrosophic Implication

| $\to_N$ | T | I | F |
|---|---|---|---|
| T | *T* | *I* | *F* |
| I | *T* | *T* | *F* |
| F | *T* | *T* | *T* |

*Table 19*

using the subjective preference that $I \rightarrow_N T$ is true (because in the classical implication $T$ is implied by anything), and $I \rightarrow_N F$ is false, while $I \rightarrow_N I$ is true because is similar to the classical implications $T \rightarrow T$ and $F \rightarrow F$, which are true.

The reader is free to check different subjective preferences.

In classical logic, one has:

### 6.9.15 Classical Equivalence

| ↔ | T | F |
|---|---|---|
| T | *T* | *F* |
| F | *F* | *T* |

*Table 20*

In neutrosophic logic, one has:

### 6.9.16 Neutrosophic Equivalence

| $\leftrightarrow_N$ | T | I | F |
|---|---|---|---|
| T | $T$ | $I$ | $F$ |
| I | $I$ | $T$ | $I$ |
| F | $F$ | $I$ | $T$ |

*Table 21*

using the subjective preference that $I \leftrightarrow_N I$ is true, because it is similar to the classical equivalences that $T \rightarrow T$ and $F \rightarrow F$ are true, and also using the prevalence:

$$I \succ_e F \succ_e T. \tag{115}$$

## 6.10 Refined Neutrosophic Literal Logic.

Each particular case has to be treated individually.

In this paper, we present a simple example.

Let's consider the following neutrosophic logical propositions:

$T$ = Tomorrow it will rain or snow.

$T$ is split into

→ Tomorrow it will rain.

→ Tomorrow it will snow.

$F$ = Tomorrow it will neither rain nor snow.

$F$ is split into

→ Tomorrow it will not rain.

→ Tomorrow it will not snow.

$I$ = Do not know if tomorrow it will be raining, nor if it will be snowing.

$I$ is split into

→ Do not know if tomorrow it will be raining or not.

→ Do not know if tomorrow it will be snowing or not.

Then:

| $\neg_N$ | $T_1$ | $T_2$ | $I_1$ | $I_2$ | $F_1$ | $F_2$ |
|---|---|---|---|---|---|---|
| | $F_1$ | $F_2$ | $T_1 \vee F_1$ | $T_2 \vee F_2$ | $T_1$ | $T_2$ |

*Table 22*

It is clear that the negation of $T_1$ (Tomorrow it will raining) is $F_1$ (Tomorrow it will not be raining). Similarly for the negation of $T_2$, which is $F_2$.

But, the negation of $I_1$ (Do not know if tomorrow it will be raining or not) is “Do know if tomorrow it will be raining or not”, which is equivalent to “We know that tomorrow it will be raining” ($T_1$), or “We know that tomorrow it will not be raining” ($F_1$). Whence, the negation of $I_1$ is $T_1 \vee F_1$, and similarly, the negation of $I_2$ is $T_2 \vee F_2$.

### 6.10.1 Refined Neutrosophic Literal Conjunction Operator

| $\wedge_N$ | $T_1$ | $T_2$ | $I_1$ | $I_2$ | $F_1$ | $F_2$ |
|---|---|---|---|---|---|---|
| $T_1$ | $T_1$ | $T_{1\,2}$ | $I_1$ | $I_2$ | $F_1$ | $F_2$ |
| $T_2$ | $T_{1\,2}$ | $T_2$ | $I_1$ | $I_2$ | $F_1$ | $F_2$ |
| $I_1$ | $I_1$ | $I_1$ | $I_1$ | $I$ | $F_1$ | $F_2$ |
| $I_2$ | $I_2$ | $I_2$ | $I$ | $I_2$ | $F_1$ | $F_2$ |
| $F_1$ | $F_1$ | $F_1$ | $F_1$ | $F_1$ | $F_1$ | $F$ |
| $F_2$ | $F_2$ | $F_2$ | $F_2$ | $F_2$ | $F$ | $F_2$ |

*Table 23*

where $T_{1\,2} = T_1 \wedge T_2 =$ “Tomorrow it will rain and it will snow”.

Of course, other prevalence orders can be studied for this particular example.

With respect to the neutrosophic conjunction, $F_l$ prevail in front of $I_k$, which prevail in front of $T_j$, or

$$F_l \succ I_k \succ T_j, \tag{116}$$

for all $l, k, j \in \{1, 2\}$.

### 6.10.2 Refined Neutrosophic Literal Disjunction Operator

| $\vee_N$ | $T_1$ | $T_2$ | $I_1$ | $I_2$ | $F_1$ | $F_2$ |
|---|---|---|---|---|---|---|
| $T_1$ | $T_1$ | $T$ | $T_1$ | $T_1$ | $T_1$ | $T_1$ |
| $T_2$ | $T$ | $T_2$ | $T_2$ | $T_2$ | $T_2$ | $T_2$ |
| $I_1$ | $T_1$ | $T_2$ | $I_1$ | $I$ | $F_1$ | $F_2$ |
| $I_2$ | $T_1$ | $T_2$ | $I$ | $I_2$ | $F_1$ | $F_2$ |
| $F_1$ | $T_1$ | $T_2$ | $F_1$ | $F_1$ | $F_1$ | $F_1 \vee F_2$ |
| $F_2$ | $T_1$ | $T_2$ | $F_2$ | $F_2$ | $F_1 \vee F_2$ | $F_2$ |

*Table 24*

with respect to the neutrosophic disjunction, $T_j$ prevail in front of $F_l$, which prevail in front of $I_k$, or

$$T_j \succ F_l \succ I_k, \qquad (117)$$

for all $j, l, k \in \{1, 2\}$.

For example, $T_1 \vee T_2 = T$ , but $F_1 \vee F_2 \notin \{T, I\ F\} \cup \{T_1, \mathrm{T}_2, I_1, \mathrm{I}_2, F_1, \mathrm{F}_2\}$.

### 6.10.3 Refinement Neutrosophic Literal Space.

**The Refinement Neutrosophic Literal Space** $\{T_1, T_2, I_1, I_2, F_1, F_2\}$ is not closed under neutrosophic negation, neutrosophic conjunction, and neutrosophic disjunction.

The reader can check the closeness under other neutrosophic literal operations.

A **neutrosophic refined literal space**

$$S_N = \{T_1, T_2, \dots, T_p;\ I_1, I_2, \dots, I_r;\ F_1, F_2, \dots, F_s\}, \qquad (118)$$

where $p, r, s$ are integers $\geq 1$, is said to be **closed** under a given neutrosophic operator "$\theta_N$", if for any elements $X, Y \in S_N$ one has $X_{\theta_N} Y \in S_N$.

Let's denote the extension of $S_N$ with respect to a single $\theta_N$ by:

$$S_{N_1}^C = (S_N, \theta_N). \qquad (119)$$

If $S_N$ is not closed with respect to the given neutrosophic operator $\theta_N$, then $S_{N_1}^C \neq S_N$ , and we extend $S_N$ by adding in the new elements resulted from the operation $X\theta_N Y$, let's denote them by $A_1, A_2, \dots A_m$.

Therefore,

$$S_{N_1}^C \neq S_N \cup \{A_1, A_2, \dots A_m\}. \qquad (120)$$

$S_{N_1}^C$ encloses $S_N$.

Similarly, we can define the **closeness** of the neutrosophic refined literal space $S_N$ with respect to the two or more neutrosophic operators $\theta_{1_N}, \theta_{2_N}, \dots, \theta_{w_N}$, for $w \geq 2$.

$S_N$ is closed under $\theta_{1_N}, \theta_{2_N}, \dots, \theta_{w_N}$ if for any $X, Y \in S_N$ and for any $i \in \{1, 2, \dots, w\}$ one has $X_{\theta_{i_N}} Y \in S_N$.

If $S_N$ is not closed under these neutrosophic operators, one can extend it as previously.

Let's consider: $S_{N_w}^C = \left(S_N, \theta_{1_N}, \theta_{2_N}, \dots, \theta_{w_N}\right)$, which is $S_N$ closed with respect to all neutrosophic operators $\theta_{1_N}, \theta_{2_N}, \dots, \theta_{w_N}$, then $S_{N_w}^C$ encloses $S_N$.

## 6.11 Conclusion.

We have defined for the first time three *neutrosophic actions* and their properties. We have introduced the *prevalence order* on $\{T, I, F\}$ with respect to a given neutrosophic operator "$o$", the *refinement of neutrosophic entities* <A>, <neutA>, and <antiA>, and the *neutrosophic literal logical operators* and the *refined literal logical operators*, and the *refinement neutrosophic literal space*.

## 6.12 References.

# 7 Neutrosophic Quadruple Numbers, Refined Neutrosophic Quadruple Numbers, Absorbance Law, and the Multiplication of Neutrosophic Quadruple Numbers

## 7.1 Abstract.

In this chapter we introduce for the first time the *neutrosophic quadruple numbers* (of the form $a + bT + cI + dF$) and the *refined neutrosophic quadruple numbers*.

Then we define an *absorbance law*, based on a *prevalence order*, both of them in order to multiply the neutrosophic components $T, I, F$ or their sub-components $T_j, I_k, F_l$ and thus to construct the *multiplication of neutrosophic quadruple numbers*.

## 7.2 Neutrosophic Quadruple Numbers.

Let's consider an entity (i.e. a number, an idea, an object, etc.) which is represented by a known part ($a$) and an unknown part ($bT + cI + dF$).

Numbers of the form:

$$NQ = a + bT + cI + dF, \quad (121)$$

where a, b, c, d are real (or complex) numbers (or intervals or in general subsets), and

$T$ = truth / membership / probability,

$I$ = indeterminacy,

$F$ = false / membership / improbability,

are called Neutrosophic Quadruple (Real respectively Complex) Numbers (or Intervals, or in general Subsets).

"$a$" is called the *known part* of *NQ*, while "$bT + cI + dF$" is called the *unknown part* of *NQ*.

## 7.3 Operations.

Let $NQ_1 = a_1 + b_1T + c_1I + d_1F, \quad (122)$

$NQ_2 = a_2 + b_2T + c_2I + d_2F, \quad (123)$

and $\alpha \in \mathbb{R}$ (or $\alpha \in \mathbb{C}$) a real (or complex) scalar.

Then:

### 7.3.1 Addition.

$$NQ_1 + NQ_2 = (a_1 + a_2) + (b_1 + b_2)T + (c_1 + c_2)I + (d_1 + d_2)F. \quad (124)$$

### 7.3.2 Substraction.

$$NQ_1 - NQ_2 = (a_1 - a_2) + (b_1 - b_2)T + (c_1 - c_2)I + (d_1 - d_2)F. \quad (125)$$

### 7.3.3 Scalar Multiplication.

$$\alpha \cdot NQ = NQ \cdot \alpha = \alpha a + \alpha bT + \alpha cI + \alpha dF. \quad (126)$$

One has:

$$0 \cdot T = 0 \cdot I = 0 \cdot F = 0, \tag{127}$$

and $mT + nT = (m+n)T,$ (128)

$$mI + nI = (m+n)I, \tag{129}$$

$$mF + nF = (m+n)F. \tag{130}$$

## 7.4 Refined Neutrosophic Quadruple Numbers.

Let us consider that Refined Neutrosophic Quadruple Numbers are numbers of the form:

$$RNQ = a + \sum_{i=1}^{p} b_i T_i + \sum_{j=1}^{r} c_j I_j + \sum_{k=1}^{s} d_k F_k, \tag{131}$$

where a, all $b_i$, all $c_j$, and all $d_k$ are real (or complex) numbers, intervals, or, in general, subsets,
while $T_1, T_2, \dots, T_p$ are refinements of $T$;
$I_1, I_2, \dots, I_r$ are refinements of $I$;
and $F_1, F_2, \dots, F_s$ are refinements of $F$.

There are cases when the known part *(a)* can be refined as well as $a_1$, $a_2$, … .

The operations are defined similarly.

Let

$$RNQ^{(u)} = a^{(u)} + \sum_{i=1}^{p} b_i^{(u)} T_i + \sum_{j=1}^{r} c_j^{(u)} I_j + \sum_{k=1}^{s} d_k^{(u)} F_k \tag{132}$$

for $u = 1$ or 2. Then:

### 7.4.1 Addition.

$$RNQ^{(1)} + RNQ^{(2)} = \left[a^{(1)} + a^{(2)}\right] + \sum_{i=1}^{p} \left[b_i^{(1)} + b_i^{(2)}\right] T_i + \sum_{j=1}^{r} \left[c_j^{(1)} + c_j^{(2)}\right] I_j + \sum_{k=1}^{s} \left[d_k^{(1)} + d_k^{(2)}\right] F_k. \quad (133)$$

### 7.4.1 Substraction.

$$RNQ^{(1)} - RNQ^{(2)} = \left[a^{(1)} - a^{(2)}\right] + \sum_{i=1}^{p} \left[b_i^{(1)} - b_i^{(2)}\right] T_i + \sum_{j=1}^{r} \left[c_j^{(1)} - c_j^{(2)}\right] I_j + \sum_{k=1}^{s} \left[d_k^{(1)} - d_k^{(2)}\right] F_k. \quad (134)$$

### 7.3.1 Scalar Multiplication.

For $\alpha \in \mathbb{R}$ (or $\alpha \in \mathbb{C}$) one has:

$$\alpha \cdot RNQ^{(1)} = \alpha \cdot a^{(1)} + \alpha \cdot \sum_{i=1}^{p} b_i^{(1)} T_i + \alpha \cdot \sum_{j=1}^{r} c_j^{(1)} I_j + \alpha \cdot \sum_{k=1}^{s} d_k^{(1)} F_k. \quad (135)$$

## 7.5 Absorbance Law.

Let $S$ be a set, endowed with a total order $x \prec y$, named "$x$ prevailed by $y$" or "$x$ less stronger than $y$" or "$x$ less preferred than $y$". We consider $x \preccurlyeq y$ as "$x$ prevailed by or equal to $y$" "$x$ less stronger than or equal to $y$", or "$x$ less preferred than or equal to $y$".

For any elements $x, y \in S$, with $x \preccurlyeq y$, one has the absorbance law:

$$x \cdot y = y \cdot x = \text{absorb}\,(x, y) = \max\{x, y\} = y, \quad (136)$$

which means that the bigger element absorbs the smaller element (the big fish eats the small fish!).

Clearly,

$$x \cdot x = x^2 = \text{absorb}\,(x, x) = \max\{x, x\} = x, \quad (137)$$

and

$$x_1 \cdot x_2 \cdot \ldots \cdot x_n = \text{absorb}(\ldots \text{absorb}(\text{absorb}(x_1, x_2), x_3) \ldots, x_n)$$
$$= \max\{\ldots \max\{\max\{x_1, x_2\}, x_3\} \ldots, x_n\}$$
$$= \max\{x_1, x_2, \ldots, x_n\}. \quad (138)$$

Analougously, we say that "$x \succ y$" and we read: "$x$ prevails to $y$" or "$x$ is stronger than $y$" or "$x$ is preferred to $y$". Also, $x \succcurlyeq y$, and we read: "$x$ prevails or is equal to $y$" "$x$ is stronger than or equal to $y$", or "$x$ is preferred or equal to $y$".

## 7.6 Multiplication of Neutrosophic Quadruple Numbers.

It depends on the prevalence order defined on $\{T, I, F\}$.

Suppose in an optimistic way the neutrosophic expert considers the prevalence order $T \succ I \succ F$. Then:

$$NQ_1 \cdot NQ_2 = (a_1 + b_1T + c_1I + d_1F) \cdot (a_2 + b_2T + c_2I + d_2F)$$
$$= a_1a_2 + (a_1b_2 + a_2b_1 + b_1b_2 + b_1c_2 + c_1b_2 + b_1d_2 + d_1b_2)T + (a_1c_2 + a_2c_1 + c_1d_2 + c_2d_1)I + (a_1d_2 + a_2d_1 + d_1d_2)F, \quad (139)$$

since $TI = IT = T, TF = FT = T, IF = FI = I$, while $T^2 = T, I^2 = I, F^2 = F$.

Suppose in an pessimistic way the neutrosophic expert considers the prevalence order $F \succ I \succ T$. Then:

$$\begin{aligned} NQ_1 \cdot NQ_2 = (&a_1 + b_1T + c_1I + d_1F) \\ &\cdot (a_2 + b_2T + c_2I + d_2F) \\ &= a_1a_2 + (a_1b_2 + a_2b_1 + b_1b_2)T \\ &+ (a_1c_2 + a_2c_1 + b_1c_2 + b_2c_1 + c_1c_2)I \\ &+ (a_1d_2 + a_2d_1 + b_1d_2 + b_2d_1 + c_1d_2 + c_2d_1 \\ &+ d_1d_2)F, \end{aligned} \tag{140}$$

since

$$F \cdot I = I \cdot F = F, F \cdot T = T \cdot F = F, I \cdot T = T \cdot I = I$$

while similarly

$$F^2 = F, I^2 = I, T^2 = T.$$

### 7.6.1 Remark.

Other prevalence orders on $\{T, I, F\}$ can be proposed, depending on the application/problem to solve, and on other conditions.

## 7.7 Multiplication of Refined Neutrosophic Quadruple Numbers

Besides a neutrosophic prevalence order defined on $\{T, I, F\}$, we also need a sub-prevalence order on $\{T_1, T_2, \dots, T_p\}$, a sub-prevalence order on $\{I_1, I_2, \dots, I_r\}$, and another sub-prevalence order on $\{F_1, F_2, \dots, F_s\}$.

We assume that, for example, if $T \succ I \succ F$, then $T_j \succ I_k \succ F_l$ for any $j \in \{1, 2, \dots, p\}$, $k \in \{1, 2, \dots, r\}$, and $l \in \{1, 2, \dots, s\}$. Therefore, any prevalence order on $\{T, I, F\}$ imposes a prevalence suborder on their corresponding refined components.

Without loss of generality, we may assume that

$$T_1 \succ T_2 \succ \cdots \succ T_p \quad (141)$$

(if this was not the case, we re-number the subcomponents in a decreasing order).

Similarly, we assume without loss of generality that:

$$I_1 \succ I_2 \succ \cdots \succ I_r \text{ , and} \quad (142)$$

$$F_1 \succ F_2 \succ \cdots \succ F_s. \quad (143)$$

### 7.7.1 Exercise for the Reader.

Let's have the neutrosophic refined space

$NS = \{T_1, T_2, T_3, I, F_1, F_2\}$,

with the prevalence order $T_1 \succ T_2 \succ T_3 \succ I \succ F_1 \succ F_2$.

Let's consider the refined neutrosophic quadruples

$NA = 2 - 3T_1 + 2T_2 + T_3 - I + 5F_1 - 3F_2$, and

$NB = 0 + T_1 - T_2 + 0 \cdot T_3 + 5I - 8F_1 + 5F_2$.

By multiplication of sub-components, the bigger absorbs the smaller. For example:

$T_2 \cdot T_3 = T_2$,

$T_1 \cdot F_1 = T_1$,

$I \cdot F_2 = I$,

$T_2 \cdot F_1 = T_2$, etc.

Multiply NA with NB.

## 7.8 References.

Symbolic (or Literal) Neutrosophic Theory is referring to the use of abstract symbols (i.e. the letters *T, I, F*, or their refined indexed letters $T_j$, $I_k$, $F_l$) in neutrosophics.

In the first chapter we extend the dialectical triad thesis-antithesis-synthesis (dynamics of <A> and <antiA>, to get a synthesis) to the neutrosophic tetrad thesis-antithesis-neutrothesis-neutrosynthesis (dynamics of <A>, <antiA>, and <neutA>, in order to get a neutrosynthesis).

In the second chapter we introduce the neutrosophic system and neutrosophic dynamic system. A neutrosophic system is a quasi- or $(t, i, f)$–classical system, in the sense that the neutrosophic system deals with quasi-terms/concepts/attributes, etc. [or $(t, i, f)$ -terms/concepts/attributes], which are approximations of the classical terms/concepts/attributes, i.e. they are partially true/membership/probable $(t\%)$, partially indeterminate $(i\%)$, and partially false/nonmembership/improbable $(f\%)$, where $t, i, f$ are subsets of the unitary interval $[0, 1]$.

In the third chapter we introduce for the first time the notions of *Neutrosophic Axiom, Neutrosophic Deducibility, Neutrosophic Axiomatic System, Degree of Contradiction (Dissimilarity) of Two Neutrosophic Axioms, etc.*

The fourth chapter we introduced for the first time a new type of structures, called *(t, i, f)-Neutrosophic Structures*, presented from a neutrosophic logic perspective, and we showed particular cases of such structures in geometry and in algebra. In any field of knowledge, each structure is composed from two parts: a *space*, and a *set of axioms* (or laws) acting (governing) on it. If the space, or at least one of its axioms (laws), has some indeterminacy of the form $(t, i, f) \neq (1, 0, 0)$, that structure is a *(t, i, f)-Neutrosophic Structure.*

In the fifth chapter we make a short history of: the *neutrosophic set, neutrosophic numerical components and neutrosophic literal components, neutrosophic numbers, etc.* The aim of this chapter is to construct examples of splitting the literal indeterminacy *(I)* into literal sub-indeterminacies $(I_1, I_2, \ldots, I_r)$, and to define a *multiplication law* of these literal sub-indeterminacies in order to be able to build *refined I-neutrosophic algebraic structures.*

In the sixth chapter we define for the first time three *neutrosophic actions* and their properties. We then introduce the *prevalence order* on $\{T, I, F\}$ with respect to a given neutrosophic operator "*o*", which may be subjective - as defined by the neutrosophic experts. And the *refinement of neutrosophic entities* <A>, <neutA>, and <antiA>. Then we extend the classical logical operators to *neutrosophic literal (symbolic) logical operators* and to *refined literal (symbolic) logical operators*, and we define the *refinement neutrosophic literal (symbolic) space*.

In the seventh chapter we introduce for the first time the *neutrosophic quadruple numbers* (of the form $a + bT + cI + dF$) and the *refined neutrosophic quadruple numbers.* Then we define an *absorbance law*, based on a *prevalence order*, both of them in order to multiply the neutrosophic components $T, I, F$ or their sub-components $T_j, I_k, F_l$ and thus to construct the *multiplication of neutrosophic quadruple numbers*.